\documentclass[10pt,twocolumn,letterpaper]{article}

\usepackage{iccv}
\usepackage{times}
\usepackage{epsfig}
\usepackage{graphicx}
\usepackage{amsmath}
\usepackage{amssymb}

\usepackage{times}
\usepackage{epsfig}
\usepackage{graphicx}
\usepackage{amsmath}
\usepackage{amssymb}
\usepackage{graphicx}
\usepackage{comment}
\usepackage{color}
\usepackage{bm}
\usepackage{amsfonts}
\usepackage{array}
\usepackage{cite}
\usepackage{booktabs}
\usepackage{float}
\usepackage{multirow}
\usepackage{mathrsfs}
\usepackage{amsfonts} 
\usepackage{lipsum}
\usepackage{wrapfig}
\usepackage[ruled]{algorithm2e}

\usepackage{color,soul}
\soulregister\ref7 
\soulregister\cite7

\usepackage[pagebackref=true,breaklinks=true,letterpaper=true,colorlinks,bookmarks=false]{hyperref}

\iccvfinalcopy 

\def\iccvPaperID{6650} 

\ificcvfinal\fi

\begin{document}

\title{RAIN: Reinforced Hybrid Attention Inference Network for Motion Forecasting}

\author{Jiachen Li$^{1,2}$\thanks{Work done during Jiachen's internship at Honda Research Institute.} ~ Fan Yang$^{3}$ ~Hengbo Ma$^{2}$ ~Srikanth Malla$^{1}$ ~Masayoshi Tomizuka$^{2}$ ~Chiho Choi$^{1}$\\
	$^{1}$Honda Research Institute USA\\
	$^{2}$University of California, Berkeley\\
	$^{3}$Carnegie Mellon University\\
	{\tt\small \{jiachen\_li, hengbo\_ma, tomizuka\}@berkeley.edu \quad fanyang3@andrew.cmu.edu} \\{\tt\small \{smalla, cchoi\}@honda-ri.com}
}

\maketitle
\ificcvfinal\thispagestyle{empty}\fi

\begin{abstract}
Motion forecasting plays a significant role in various domains (e.g., autonomous driving, human-robot interaction), which aims to predict future motion sequences given a set of historical observations.
However, the observed elements may be of different levels of importance. 
Some information may be irrelevant or even distracting to the forecasting in certain situations. 
To address this issue, we propose a generic motion forecasting framework (named RAIN) with dynamic key information selection and ranking based on a hybrid attention mechanism.
The general framework is instantiated to handle multi-agent trajectory prediction and human motion forecasting tasks, respectively.
In the former task, the model learns to recognize the relations between agents with a graph representation and to determine their relative significance.
In the latter task, the model learns to capture the temporal proximity and dependency in long-term human motions.
We also propose an effective double-stage training pipeline with an alternating training strategy to optimize the parameters in different modules of the framework.
We validate the framework on both synthetic simulations and motion forecasting benchmarks in different domains, demonstrating that our method not only achieves state-of-the-art forecasting performance, but also provides interpretable and reasonable hybrid attention weights.
\end{abstract}

\vspace{-0.3cm}
\section{Introduction}

Motion forecasting has been widely studied in various domains, such as physical systems, human skeletons, and multi-agent interacting systems (e.g., traffic participants, sports players, etc).
The problem is formulated as to predict future states or trajectories based on historical spatio-temporal observations.
However, the observed information may be of different levels of significance and in some situations not all the information is relevant for the forecasting. 
Moreover, the key information may be varying as the situation evolves, which motivates the forecasting approach to dynamically adjust its attention to different subsets of observations.
Here we provide two illustrative real-world examples where key information is naturally selected based on either spatial relations or temporal dependencies.
An on-road vehicle usually only needs to pay attention to the traffic participants that are interacting or having a conflict with itself, so only a subset of observations are indeed relevant when predicting its future behavior.
For human motion forecasting, it is observed that humans tend to repeat their motions, which motivates dynamic attention to different segments of previous motions given the current observation.

\begin{figure}[!tbp]
	\centering
	\includegraphics[width=0.95\columnwidth]{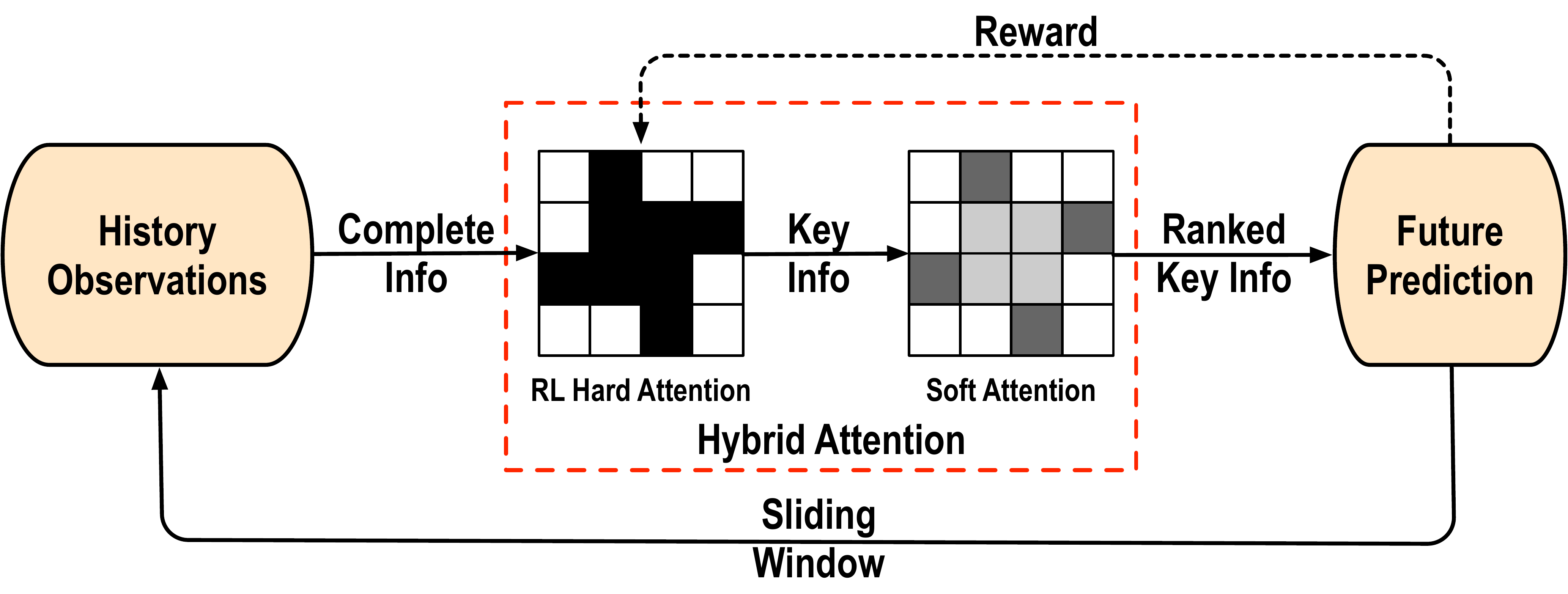}
	\caption{A high-level diagram of the proposed general prediction framework with key information/element selection and ranking, which consists of two major parts: an RL-based hard attention mechanism to discriminate key information from complete observations and a soft attention mechanism to further figure out relative significance of the key information. The whole procedure can be iteratively applied over time with a sliding window to enable dynamic selection of key information to adapt to evolving situations.}
	\vspace{-0.3cm}
	\label{fig:framework}
\end{figure}

Attention mechanisms have been widely adopted to learn the relative importance of elements.
There are two major types of attention mechanisms in literature: soft attention and hard attention \cite{xu2015show}.
The soft attention is usually performed by applying a score function to input features followed by a $\mathrm{softmax}$ function to obtain the attention weights in the range of $[0, 1]$. These operations are fully differentiable which can be trained by back-propagation with typical gradient-based optimizers. 
However, the $\mathrm{softmax}$ function tends to assign non-zero attention weights to irrelevant or unimportant elements, which dilutes the attention given to the truly significant information \cite{vemula2018social,shen2018reinforced}.
In contrast, the hard attention mechanism can force the model to only pay attention to the relevant information while discard the others entirely to reduce information redundancy.
The hard attention weights can be only binary: 0 (discarded) or 1 (retained).
However, the hard attention is not differentiable due to the $\mathrm{argmax}$ operation, which needs to be optimized by reinforcement learning (RL) algorithms (e.g., deep Q-learning \cite{mnih2015human}, policy gradient \cite{williams1992simple}).
Recently, some alternatives to the traditional hard attention have been proposed based on the approximation of evidence lower bound (ELBO) \cite{maddison2016concrete}, which can be trained end-to-end.

The general idea of selecting the most important information/elements with hard attention has been applied to several different domains, such as computer vision and natural language processing. 
Wu et al \cite{wu2019multi} introduced a key frame selection framework based on multi-agent reinforcement learning for video based human activity recognition.
Wang et al \cite{wang2020active} presented a framework for informative view selection from multiple indoor cameras to recognize human actions.
Gao et al \cite{gao2019graph} proposed a hard and channel-wise attention network for graph representation learning.
Shen et al \cite{shen2018reinforced} illustrated a reinforced self-attention network to figure out sparse dependencies between tokens in a sentence.

However, the efficacy of hard attention in motion forecasting tasks, to the best of our knowledge, still remains largely unexplored so far.
Besides, many existing works including the aforementioned ones, pre-define a fixed number of elements to pay attention to, which may be unsatisfactory in the scenarios where the amount of key information/elements is varying.
For example, the motion of a certain entity in an interacting system may be affected by a varying number of entities at different time, thus a fixed number of selected elements may be redundant or insufficient in different situations.
To address this issue, we propose a reinforcement learning based hard attention mechanism for motion forecasting, which does not enforce any constraints on the amount of key elements. It is even possible in some situations that no element or all the elements are selected based on their significance. 
In the multi-agent setting with a graph representation, learning hard attention can also be treated and interpreted as graph structure/topology learning \cite{pmlr-v97-franceschi19a,li2020causal}.
Since the selected key information may be still at different levels of importance, we propose to employ soft attention as a ranking mechanism to further discriminate relative importance.

To the best of our knowledge, we are the first to propose a hybrid attention based framework for motion forecasting, which is illustrated in Figure \ref{fig:framework}. 
The main contributions of this paper are summarized as:

$\bullet$ We propose a general motion forecasting framework (named RAIN) with dynamic key information/element selection and ranking via a hybrid attention mechanism.

$\bullet$ We propose an effective double-stage training pipeline with an alternating training strategy to improve different modules in the framework alternatively.

$\bullet$ We instantiate the general framework and propose a novel graph-based model for multi-agent trajectory forecasting. 
We also demonstrate the general idea on human skeleton motion forecasting, where a state-of-the-art model \cite{wei2020his} is employed as a part of our framework.
We validate the proposed framework on both domains and our method achieves the state-of-the-art performance consistently.

\begin{figure*}[!tbp]
	\centering
	\includegraphics[width=\textwidth]{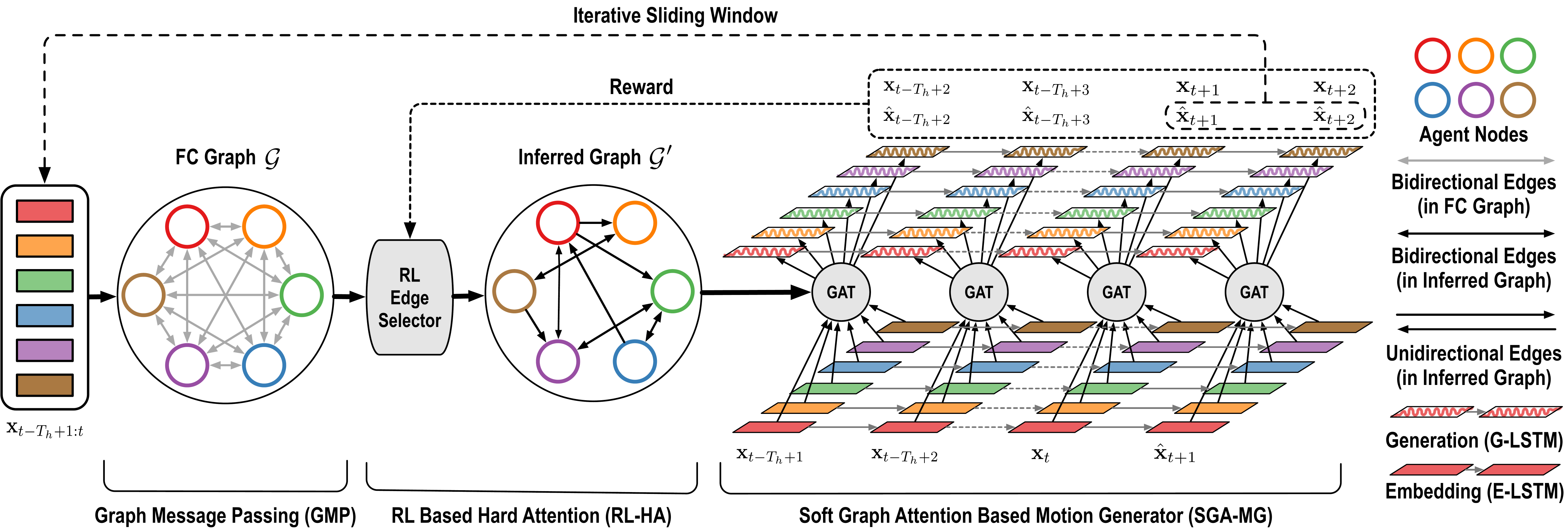}
	\caption{The diagram of the proposed hybrid attention based prediction framework for multi-agent interacting systems, which consists of three major components: a graph message passing module (GMP), a RL based hard attention module (RL-HA) and a soft graph attention based motion generator (SGA-MG). 
		GMP aims to pass the information across the fully connected graph $\mathcal{G}$.
		RL-HA aims to figure out truly relevant information (edges) and output an inferred relation graph $\mathcal{G}'$, while SGA-MG aims to further rank the significance of the key information with soft attention and generate future motions with recurrent neural networks. The prediction process can be either one-shot (with static relation graph $\mathcal{G}'$) or progressive with a sliding window (with dynamic relation graph $\mathcal{G}'$). Best viewed in color.}
	\label{fig:model_diagram}
\end{figure*}

\section{Related Work}

\subsection{Trajectory/Motion Forecasting}
Many research efforts have been devoted to motion forecasting in various domains. Here we particularly provide a brief review of literature on physical systems, highly interactive traffic scenarios and skeleton based human motions, which are closely related to this work.\\
$\bullet$ \textbf{Physical systems}: 
Learning the dynamics of physical systems involving multiple interacting elements have been studied in some recent works, either from simulated trajectories \cite{battaglia2016interaction,hoshen2017vain,kipf2018neural,li2020evolvegraph} or from generated videos \cite{watters2017visual,van2018relational,Kossen2020Structured}, where graph neural network is used to model interactions. 
Some of them assume a known graph topology based on prior knowledge, while the others infer the underlying structure explicitly or implicitly. 
Our approach falls into the second category, where we propose a novel fashion of relational inference based on reinforcement learning.
\\
$\bullet$ \textbf{Traffic scenarios}: 
A number of works attempt to forecast future trajectories or behaviors of heterogeneous traffic participants (e.g., pedestrians, vehicles, cyclists) \cite{ma2021continual,huang2019stgat,gang2021loki,kosaraju2019social,wu2020motionnet,li2019conditional,phan2020covernet,ma2020reinforcement,choi2019looking,chai2020multipath,ma2021multi}. In order to model the relations or interactions between different entities, many information aggregation techniques have been developed, including but not limited to, feature concatenation \cite{li2019conditional}, attention mechanisms \cite{vemula2018social,li2021spatio,huang2019stgat,kosaraju2019social}, social pooling \cite{alahi2016social,deo2018convolutional}, message passing across graphs \cite{li2020evolvegraph,huang2019stgat,choi2021shared}, etc.
Moreover, recent works are also putting more emphasis on environmental modeling to leverage physical and semantic constraints such as road layout and traffic rules in driving scenarios \cite{gao2020vectornet,tang2019multiple,phan2020covernet}.
\\
$\bullet$ \textbf{Human motions}: 
Recurrent neural networks are widely utilized in human motion forecasting \cite{jain2016structural,gopalakrishnan2019neural,fragkiadaki2015recurrent}.
Recent works also proposed to adopt feed-forward methods for effective encoding of long-term motion history \cite{mao2019learning,li2018convolutional}.
Besides, attention mechanisms were also employed due to its flexibility and efficacy.
In \cite{tang2018long}, Tang et al adopted a frame-wise attention mechanism to summarize the pose history. 
In \cite{wei2020his}, Mao et al proposed a motion attention based method, which achieved the previous state-of-the-art performance. 
In this paper, we apply our hard attention module on top of their soft attention model \cite{wei2020his} to illustrate the efficacy of our proposed hybrid attention based framework.

\subsection{Attention Mechanisms}
Attention mechanisms have been widely used in deep neural networks due to their efficacy and efficiency for sequence modeling and information fusion. Existing works have demonstrated the superiority of attention in a broad range of domains, such as natural language processing \cite{parikh2016decomposable,shen2018reinforced,wang2016attention}, image captioning \cite{xu2015show,wang2017residual,anderson2018bottom}, saliency detection \cite{liu2018picanet,zhao2019pyramid}, time-series modeling \cite{fan2019multi,qin2017dual}, human activity recognition \cite{si2019attention,cho2020self,song2018spatio}.
Most of these papers either solely apply soft attention to obtain contextual embeddings or solely apply hard attention to select important elements.
The only exception is \cite{shen2018reinforced} in which a hybrid attention was presented to figure out sparse dependencies between sequential tokens in a sentence. 
However, their method can only be applied to sequence modeling, which is hard to generalize to the motion forecasting task of multi-agent interacting systems.
For the motion forecasting task, some existing works only adopt soft attention to weight the complete observed information \cite{vemula2018social,li2019conditional,huang2019stgat,li2021spatio,kosaraju2019social,wei2020his}, while our method can discriminate the most relevant elements via the hybrid attention.

\section{Problem Overview}
Since the proposed approach can be widely applied to various domains, we introduce the problem formulation in a general way.
Define a multivariate dynamic system
\vspace{-0.1cm}
\begin{equation}
	\mathbf{X}_{t+1:t+T_f} = f(\mathbf{X}_{t-T_h+1:t}, \mathbf{C}),
	\vspace{-0.2cm}
\end{equation}
where $\mathbf{X}_t = \{\mathbf{x}^i_t, i=1,...,N\}$ denotes the system state at time $t$ and $\mathbf{C}=\{\mathbf{c}^i, i=1,...,N\}$ denotes optional context information or external factors. 
$N$ is the total number of variables which have a specific meaning in different domains.
The goal of this work is to approximate the conditional distribution $p(\mathbf{X}_{t+1:t+T_f}|\mathbf{X}_{t-T_h+1:t}, \mathbf{C})$, where $T_h$ and $T_f$ denote the history and prediction horizon.

For a multi-agent interacting system, the variables refer to the involved homogeneous/heterogeneous interacting entities, where the state may include position, velocity, etc.
For a multivariate time series such as human motions, the variables refer to a set of human skeletons, where the state may include joint coordinates or relative angles.

In this paper, we instantiate and apply the proposed general prediction framework illustrated in Figure \ref{fig:framework} to both multi-agent systems with spatially interacting agents (e.g., physical systems, traffic participants), and multivariate time series with temporal dependency (e.g., skeleton-based human motions).
For long-term prediction of multi-agent interacting systems, we propose a novel, complete model architecture based on graph representation in Section \ref{sec:multi-agent-method}.
For multivariate time series (i.e., human motions), we present an effective way to build our proposed hard attention module on top of a state-of-the-art soft attention based model for human motion prediction \cite{wei2020his} in Section \ref{sec:human-motion-method}.

\section{RAIN for Multi-Agent Interacting System}\label{sec:multi-agent-method}

\subsection{Framework Overview}

An illustrative diagram is shown in Figure \ref{fig:model_diagram} to introduce the pipeline and three major components: graph message passing module (GMP), RL based hard attention module (RL-HA) and soft graph attention based motion generator (SGA-MG), which cooperate closely to improve the final prediction performance. 
More specifically, for the prediction of a certain target entity, GMP collects information from other entities across graph $\mathcal{G}$. 
RL-HA discriminates the key relevant elements from the complete observations and provides SGA-MG with an inferred relation graph $\mathcal{G}'$ with only selected edges, which is a natural generalization of the traditional hard attention to graph representation. 
SGA-MG uses soft attention weights to rank the relative importance of key information and generates future trajectories. 
The prediction together with the ground truth provides rewards to RL-HA during the training phase to guide the improvement of the RL edge selector.
GMP is pre-trained to collect contextual information across the whole graph.
SGA-MG is pre-trained with a fully connected topology in order to improve training efficiency and stability as well as to enable informative initial reward.

\subsection{Graph Message Passing}\label{sec:GMP}
It is natural to represent a multi-agent system with $N$ entities as a fully connected (FC) graph $\mathcal{G}=(\mathcal{V},\mathcal{E})$, where $\mathcal{V}=\{\mathbf{v}_i,i=1,...,N\}$ and $\mathcal{E}=\{\mathbf{e}_{ij},i,j=1,...,N\}$. $\mathbf{v}_i$ denotes node $i$'s attribute and $\mathbf{e}_{ij}$ denotes the edge attribute from sender node $j$ to receiver node $i$.
The node attribute consists of a self-attribute to store the individual information, a social-attribute to store other entities' information, and a context-attribute to include the agent's context information. 
More formally, we have
$\mathbf{v}^{\text{self}}_i = f^m_s(\mathbf{x}^i_{t-T_h+1:t})$,  $\mathbf{v}^{\text{neighbor}}_i = f^m_n(\mathbf{x}^i_{t-T_h+1:t})$,
$\mathbf{v}^{\text{context}}_i = f_c(\mathbf{c}^i)$,
where $m\in \{1,...M\}$, $M$ is number of agent types. $f^m_s$ and $f^m_n$ are state embedding functions, and $f_c$ is context embedding function. Different state embedding functions corresponding to certain agent types are applied to heterogeneous agents.
 
Since the relationship between a pair of agents is not only determined by their own behaviors but also affected by other agents in the scene, it is not sufficient to use only a pair of self node attributes to determine the existence of an edge.
Therefore, we apply a round of message passing to collect the contextual information across the FC graph $\mathcal{G}$ by
\begin{align}
	\alpha_{ij} &= \frac{\exp \left(\text{MLP}\left(\left[\mathbf{v}^{\text{self}}_i || \mathbf{v}^{\text{neighbor}}_j\right]\right)\right)}{\sum\nolimits_{k \in \mathcal{N}_i} \exp \left(\text{MLP}\left(\left[\mathbf{v}^{\text{self}}_i || \mathbf{v}^{\text{neighbor}}_k\right]\right)\right)}, \\
	&\mathbf{v}^{\text{social}}_i = f_v\left(\sum\nolimits_{j \in \mathcal{N}_i} \alpha_{ij} \mathbf{v}^{\text{neighbor}}_j\right),
\end{align}
where $f_v$ is the social attribute update function and $||$ denotes the concatenation operation. $\mathcal{N}_i$ denotes the set of one-hop neighbors of node $i$. MLP refers to multi-layer perceptron.
The complete node attribute is 
\begin{align}
	\mathbf{v}_i = f_{\text{enc}}\left(\left[\mathbf{v}^{\text{self}}_i,\ \mathbf{v}^{\text{social}}_i,\ \mathbf{v}^{\text{context}}_i\right]\right).
\end{align}

\subsection{Hard Attention: Key Information Selection}

As shown in Figure \ref{fig:framework}, the RL-based hard attention module serves as a key information selector, which takes the complete history observations as input and discriminates the truly relevant information while totally discard the rest.
More specifically, in the context of multi-agent system, the RL-HA module is expected to figure out truly influencing factors when predicting the motions of a certain agent. In other words, the goal is to assert the existence of each edge in the FC graph based on the observations so that the redundant information is discarded in the prediction.

The selection of key edges naturally fits into a reinforcement learning framework.
The definition of observations, actions and reward functions are elaborated in the following, while the training procedures are left to Section \ref{sec:training}.

\textbf{Observations:}
The observation $O$ of RL-agent at RL-step $\eta \ (\leq T_{\text{RL}})$ includes a pair of node attributes $\mathbf{v}_i$ and $\mathbf{v}_j$ as well as the current edge selection status $s_{ij}$ (0: ``retained" or 1: ``discarded"). $T_{\text{RL}}$ is the upper bound of RL-steps.
The observation $O_\eta$ is obtained by $O_\eta = [\mathbf{v}_i,\ \mathbf{v}_j,\ s_{ij,\eta}]$.
Note that the dimension of $O_\eta$ only depends on the dimension of node attributes, which enables the applicability to the systems with varying numbers of entities.
The policy network of RL-agent takes the observation $O_{\eta}$ as input and decides the action at the current RL-step.

\textbf{Actions:}
There are two possible actions for the RL-agent: ``staying the same" (action 0) and ``changing to the opposite" (action 1).
At each RL-step, the RL-agent makes decision for each edge in the FC graph. 
The policy can be written as $a=\pi(O)$.
We do not enforce any constraints on the selection of edges, i.e., there is no lower/upper bound on the number of selected edges.
The actions of RL-agent may change the topology of the inferred graph $\mathcal{G}'$ after each RL-step, which further influence the SGA-MG module.

\textbf{Rewards:}
In general, the reward indicates how good the action taken by the RL-agent is with respect to the current situation.
In our method, the reward is designed to indicate the performance of motion forecasting in various aspects. The acquisition of high rewards depends on the collaboration of all the modules in the framework.

The reward consists of three parts: regular reward $R_\text{reg}$, improvement reward $R_\text{imp}$ and stimulation/punishment $R_\text{sti}/R_\text{pun}$.
More specifically, the \textit{regular reward} is the negative mean squared error of future predictions calculated by
\begin{align}
	R_{\text{reg},\eta} = - \frac{1}{N} \sum_{i=1}^{N} \sum_{t'=t+1}^{t+T_f}||\mathbf{x}^i_{t'} - \hat{\mathbf{x}}^i_{t',\eta}||^2.
\end{align}
The \textit{improvement reward} encourages the decrease of prediction error via applying a sign function to the error change between consecutive RL-steps, which is obtained by
\begin{align}
	R_{\text{imp},\eta} =  \mathrm{sign}(R_{\text{reg},\eta} - R_{\text{reg},\eta-1}).
\end{align}
The reason of applying a $\mathrm{sign}$ function instead of directly using the raw improvement is to avoid reward vanishing when improvement becomes smaller towards convergence.
The \textit{stimulation/punishment} is applied when there is a large improvement or deterioration in terms of a certain metric, which is given by
\begin{align}
	R_{\text{sti},\eta} = \Omega_s, \quad R_{\text{pun},\eta} = -\Omega_p,
\end{align}
where $\Omega_s$ and $\Omega_p$ are manually defined positive constants.
These rewards depend on the metrics in specific domains.

Then the whole reward is calculated by
\vspace{-0.2cm}
\begin{align}
	R_{\eta} = \ &R_{\text{reg},\eta} + \beta_\text{imp} R_{\text{imp},\eta} \nonumber\\
	&+ \beta_\text{sti} R_{\text{sti},\eta} \mathbb{I}(\text{sti}) + \beta_\text{pun} R_{\text{pun},\eta} \mathbb{I}(\text{pun}),
\end{align}
where $\beta_\text{imp}$, $\beta_\text{sti}$ and $\beta_\text{pun}$ are hyperparameters and $\mathbb{I}(\cdot)$ is an indicator function to indicate the occurrence of large improvement or deterioration.

\subsection{Soft Attention: Key Information Ranking}

After finalizing the key information (edges) by the RL-HA module, the soft graph attention mechanism \cite{vel2018graph} is applied over the inferred graph $\mathcal{G}'$ to further determine the relative significance of the selected key information at each time step. 
Here we take time step $t$ as an example to illustrate the soft graph attention mechanism.
Note that unlike the one-shot state embedding in GMP, here we construct a spatio-temporal graph to incorporate the state information with a stepwise embedding strategy.

In order to avoid confusion on notation with the RL-HA section, here we denote $i$-th node attribute at time $t$ as 
\begin{align}
	\bar{\mathbf{v}}_{i,t}=[\bar{\mathbf{v}}^{\text{self}}_{i,t},\ \bar{\mathbf{v}}^{\text{social}}_{i,t},\ {\mathbf{v}}^{\text{context}}_{i}],
\end{align}
where the context attribute ${\mathbf{v}}^{\text{context}}_{i}$ is obtained in Section \ref{sec:GMP}, the self-attribute $\bar{\mathbf{v}}^{\text{self}}_{i,t}$ and the intermediate attribute $\bar{\mathbf{v}}^{\text{neighbor}}_{i,t}$ are introduced in Section \ref{sec:MG}, and the social attribute $\bar{\mathbf{v}}^{\text{social}}_{i,t}$ is calculated by the graph soft attention mechanism as follows
\begin{equation}
	\begin{aligned}
		\bar{\alpha}^t_{ij} =& \frac{\exp\left(\text{MLP}\left(\left[\bar{\mathbf{v}}^{\text{self}}_{i,t} || \bar{\mathbf{v}}^{\text{neighbor}}_{j,t}\right]\right)\right)}{\sum_{k\in \mathcal{N}_i}\exp\left(\text{MLP}\left(\left[\bar{\mathbf{v}}^{\text{self}}_{i,t}|| \bar{\mathbf{v}}^{\text{neighbor}}_{k,t}\right]\right)\right)}, \\
		&\bar{\mathbf{v}}^{\text{social}}_{i,t} = \bar{f}_v \left(\sum\nolimits_{j\in \mathcal{N}_i} \bar{\alpha}^t_{ij}\bar{\mathbf{v}}^{\text{neighbor}}_{j,t}\right), 
	\end{aligned}
\end{equation}
where $\bar{\alpha}^t_{ij}$ are learnable attention weights.
We also use multi-head attention to stabilize training \cite{vel2018graph}.

\subsection{Spatio-Temporal Motion Generator}\label{sec:MG}
The motion generator consists of two LSTM networks (E-LSTM/G-LSTM) with the soft graph attention in between. 
The E-LSTM takes in agent state information and outputs $\bar{\mathbf{v}}^{\text{self}}_{i,t}$ at each time step, while the G-LSTM takes in the complete node attribute $\bar{\mathbf{v}}_{i,t}$ and outputs the predicted change in state $\Delta \hat{\mathbf{x}}^i_t$ at the current time $t$, which is used to calculate the state $\hat{\mathbf{x}}^i_{t+1}$ with the system model (e.g., discrete-time linear dynamics).
More specifically at time $t$,
\begin{align}
	\text{Embedding:}& \ \bar{\mathbf{v}}^{\text{self}}_{i,t} = \text{E-LSTM}^\text{s}(\mathbf{x}^i_{t}; \mathbf{h}^{s,i}_{t}), \\
	&  \bar{\mathbf{v}}^{\text{neighbor}}_{i,t}  = 
	\text{E-LSTM}^\text{n}(\mathbf{x}^i_{t}; \mathbf{h}^{n,i}_{t}), \\
	\text{Generation:}& \ \Delta \hat{\mathbf{x}}^i_{t} = \text{G-LSTM}(\bar{\mathbf{v}}_{i,t}; \tilde{\mathbf{h}}^i_{t}),\\
	& \ \hat{\mathbf{x}}^i_{t+1} = f_\text{system}(\mathbf{x}^i_{t}, \Delta\hat{\mathbf{x}}^i_{t})+ \bm{\epsilon},
\end{align} 
where $\mathbf{h}^{s,i}_{t}$, $\mathbf{h}^{n,i}_{t}$, and $\tilde{\mathbf{h}}^i_t$ are the hidden states of E-LSTM and G-LSTM respectively, $\bm{\epsilon} \sim \mathcal{N}(\bm{0}, \bm{\Sigma})$ is a random noise to incorporate uncertainty, and $f_\text{system}$ is the system model.
Note that the whole generation process is divided into two stages: burn-in stage (from $t-T_h+1$ to $t$) and prediction stage (from $t+1$ to $t+T_f$). At burn-in stage, the true state is fed into E-LSTM while at prediction stage, the last prediction is fed instead.

If the topology of the inferred graph $\mathcal{G}'$ is assumed to remain static over time, we can use one-shot generation to obtain the complete future trajectory. 
Otherwise, we can first generate the trajectory segment within a certain future horizon $\tau < T_f$, and push the predicted segment into the observations. This process can be iteratively propagated to generate the whole trajectory. We set $\tau=2$ in Figure \ref{fig:model_diagram}.

\subsection{Framework Training Strategy}\label{sec:training}

Since the RL-HA module is not differentiable, we cannot use an end-to-end fashion to train the whole model.
Therefore, we propose an alternating training algorithm to improve RL-HA and the other modules separately.
There are two training stages: pre-training stage and formal-training stage.
In the \textit{pre-training} stage, we pre-train the parameters of GMP module with an auto-encoder structure by unsupervised learning, where GMP serves as the encoder together with an auxiliary decoder. The decoder is discarded after the GMP is well trained.
In addition, we pre-train the SGA-MG module with a fully connected topology to enable informative initial reward for the RL-HA module.
The pre-training is necessary for the convergence of formal-training stage.
In the \textit{formal-training} stage, the pre-trained parameters of GMP are fixed to serve as a feature extractor, and we perform alternating optimization of RL-HA and SGA-MG modules: (a) train the RL-HA module with fixed parameters of GMP and SGA-MG with Double Deep Q-Learning (DDQN) method \cite{hasselt2016deep}; (b) finetune SGA-MG with fixed parameters of GMP and RL-HA using back-propagation methods.
Please refer to Section \ref{sec:training_alg} in the supplementary materials for the pseudo-code and implementation details.

\section{RAIN for Human Skeleton Motions}\label{sec:human-motion-method}

In this section, we address the application of our method to capture long-term temporal dependency in skeleton based human motions.
Instead of selecting important or relevant entities introduced in Section \ref{sec:multi-agent-method}, here we propose to utilize the RL-HA module to discriminate key information over the whole history horizon. It can be either frame-wise selection or segment-wise selection. We demonstrate an illustrative case study of the latter based on a state-of-the-art model proposed in \cite{wei2020his}.

More specifically, the model in \cite{wei2020his} is employed as the soft attention based motion generator in our framework.
The proposed RL hard attention mechanism is built on top of it, and they work as a whole to generate future human motions.
First, an encoding function is applied to extract contextual representation of each history frame.
Then, the RL-HA module discriminates the key motion segments from history observations, and the soft attention mechanism further outputs relative significance of the selected segments for the current prediction.
The model architecture and implementation details are elaborated in Section \ref{sec:RAIN_human}.

\begin{figure}
	\centering
	\includegraphics[width=\columnwidth]{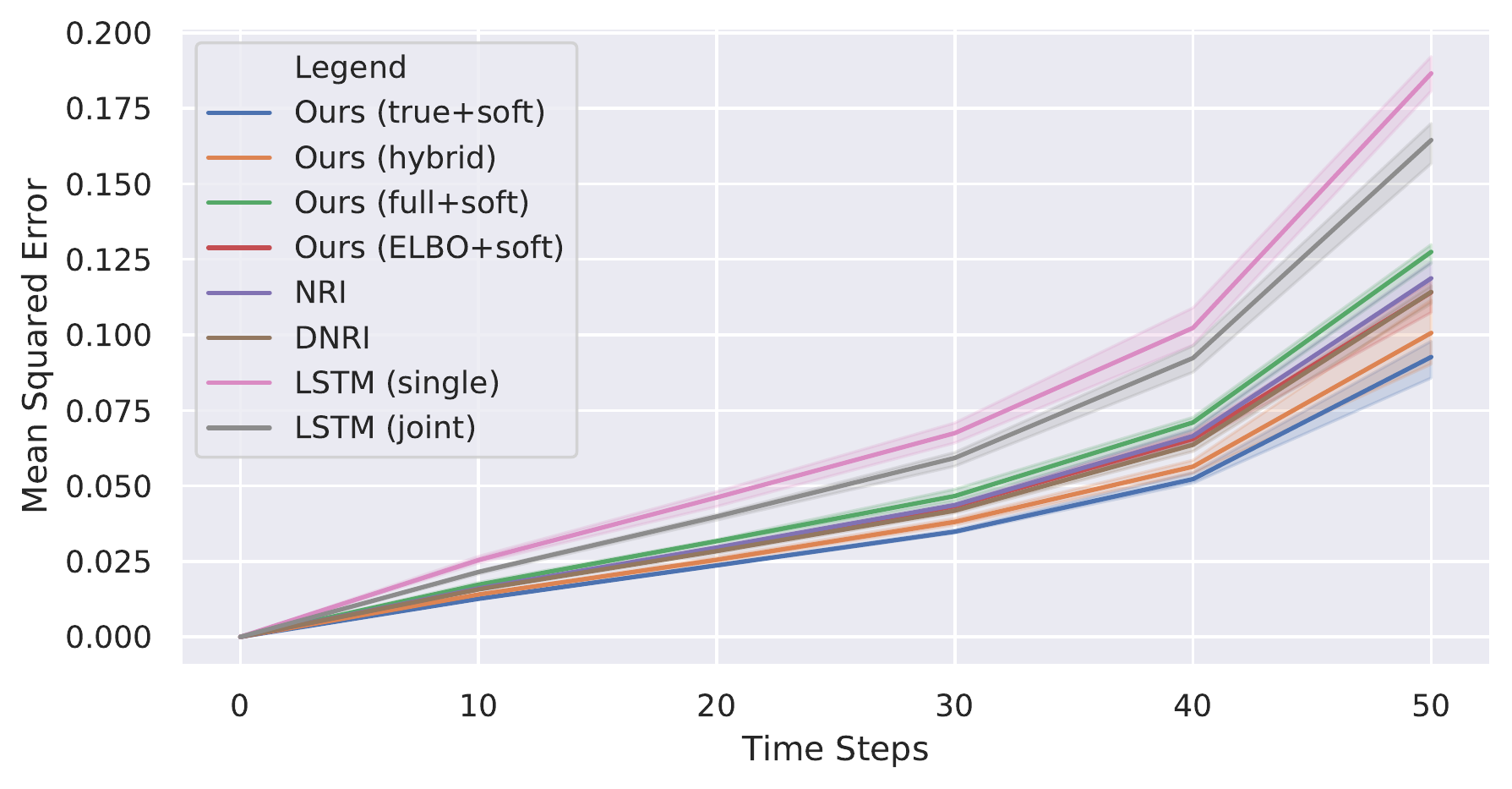}
	\caption{Mean squared error (MSE) in predicting future positions of mixed particles.}
	\vspace{-0.2cm}
	\label{fig:particle_error}
\end{figure}

\section{Experiments}\label{sec:exp}
In this paper, we validated the proposed hybrid attention approach on the datasets in different domains, including a synthetic physics motion dataset, a real-world driving dataset (nuScenes \cite{nuscenes2019}) and a human motion dataset (Human3.6M \cite{h36m_pami}). 

For the mixed particle simulation, since we have access to the ground truth of the particle charging state (i.e., positively/negatively-charged or uncharged), we can quantitatively and qualitatively evaluate the model performance in terms of both attention learning and mean square error (MSE) of particle position prediction.
For the nuScenes driving dataset, we employ three widely used metrics: minimum average displacement error (minADE$_{20}$), minimum final displacement error (minFDE$_{20}$) and miss rate (MR) in trajectory prediction \cite{lee2017desire}. We report the results for vehicles and pedestrians separately.
For the Human3.6M motion dataset, we adopt the standard metrics in the human pose estimation and motion forecasting literature \cite{wei2020his}, which is the Euclidean distance in Euler angle representation.
We also provide ablative analysis on these datasets.

In all the experiments, we used a batch size of 32 and the Adam optimizer with an initial learning rate of 0.001 to train the models with a single NVIDIA Quadro RTX 6000 GPU. 
The additional experimental results, dataset details, baseline methods as well as implementation details are elaborated in supplementary materials.

\subsection{Synthetic Simulation: Mixed Particle System}

\begin{figure*}
	\centering
	\includegraphics[width=\textwidth]{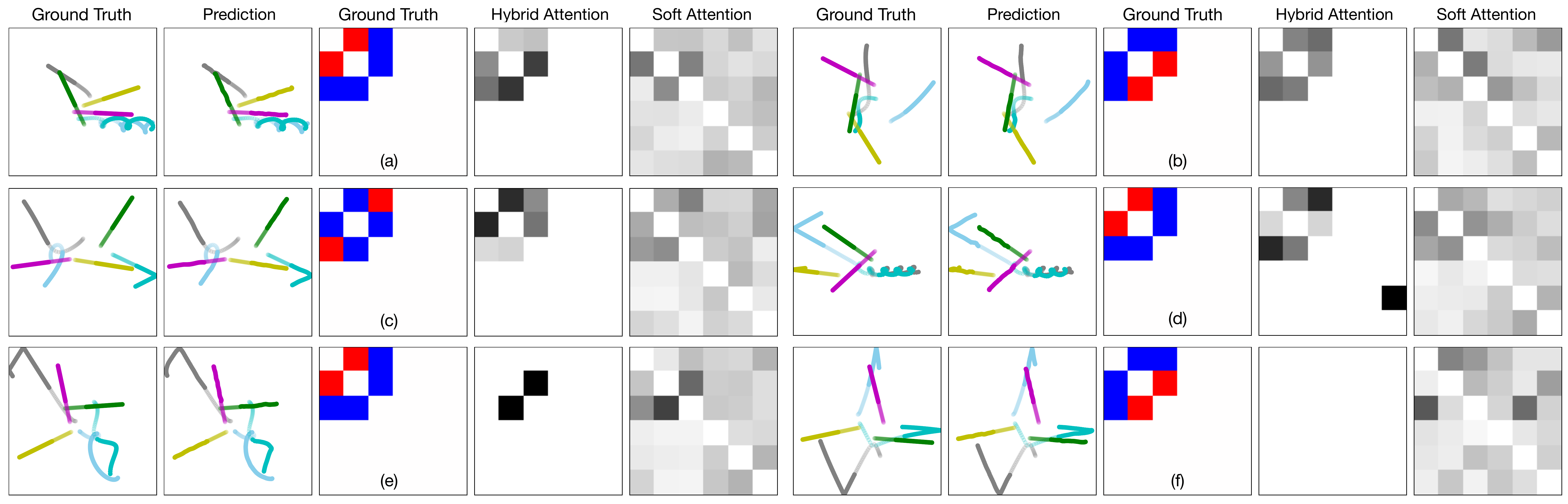}
	\caption{Visualization of particle trajectories and attention maps. The semi-transparent segments represent the observation of 30 time steps and the solid ones represent the prediction of future 50 time steps. In the ground truth relation map, red denotes ``repel" and blue denotes ``attract". In the attention maps, darker color indicates larger weight. Note that we do not consider self-attention (i.e. zero diagonal).}
	\vspace{-0.3cm}
	\label{fig:particle_plot}
\end{figure*}

\begin{table}[!tbp]
	\centering
	\resizebox{\columnwidth}{!}{
		\begin{tabular}{l|cccc}
			\hline
			& Accuracy & Precision & Recall & F1-score   \\
			\hline 
			\hline
			Corr. LSTM \cite{kipf2018neural} & 71.85$\pm$0.39 & 54.52$\pm$0.24 & 60.17$\pm$0.16 & 57.21$\pm$0.33  \\
			NRI \cite{kipf2018neural} & 85.47$\pm$0.98 & 78.57$\pm$0.59 & 64.71$\pm$0.66 & 70.97$\pm$0.51    \\
			DNRI \cite{graber2020dynamic} & 87.54$\pm$1.66 & 82.48$\pm$0.91 & 67.23$\pm$0.78 & 74.08$\pm$0.83 \\
			\hline
			Ours (ELBO+soft) & 89.31$\pm$0.46 & 87.29$\pm$0.21 & 73.52$\pm$0.30 & 79.82$\pm$0.25 \\
			Ours (hybrid) & \textbf{93.52}$\pm$1.25 & \textbf{92.34}$\pm$0.87 & \textbf{80.55}$\pm$0.92 & \textbf{86.04}$\pm$0.74 \\
			\hline
			Supervised & 97.48$\pm$0.23 & 95.24$\pm$0.09 & 89.55$\pm$0.14 & 92.31$\pm$0.11  \\
			\hline
	\end{tabular}}
	\vspace{0.cm}
	\caption{Evaluation (Mean$\pm$Std in \%) of Relation Recognition (Mixed Particle System).}
	\vspace{-0.cm}
	\label{tab:edge_recovery}
\end{table}

We applied our approach to a simulated physical system with mixed charged and uncharged particles, where the charged ones are randomly assigned positive or negative charges with a uniform distribution. 
The particles with the same charge repel and the ones with opposite charges attract according to the fundamental law of electricity. The uncharged particles perform uniform motion independently.
The proposed hybrid-attention based prediction model is expected to learn the relations between particles and forecast their long-term motions.

The goal of RL-based hard attention mechanism is to infer whether there is a force between a pair of particles, which is essentially a binary classification task. The recognition of the force type (repel or attract) will be handled implicitly by the soft attention counterpart.
The relation recognition results of different approaches are compared in Table \ref{tab:edge_recovery}, where the standard deviations are calculated based on three independent runs.
It shows that the supervised learning method which directly trains a binary classifier with ground truth labels performs the best in terms of all the evaluation metrics. However, the accessibility of true relation labels is highly limited in real-world tasks.
Among the approaches that do not require true labels during training, our method achieves the highest recognition accuracy, which improves by 6.8\% over the strongest baseline DNRI and 4.7\% over the ablative baseline \textit{Ours (ELBO+soft)}. Based on the experiments, end-to-end training with ELBO tends to be more stable and converges faster than RL in general, but it did not achieve a better final performance.
The accuracy alone cannot sufficiently prove the superiority due to the data imbalance of the two classes. 
Therefore, we also compared the precision, recall and F1-score.
Our method achieves consistently better results on these metrics.

We further predicted the particle positions at the future 50 time steps given the historical observations of 30 time steps.
The comparison of mean squared error (MSE) of different methods is provided in Figure \ref{fig:particle_error}. The solid lines and shaded areas represent the average value and standard deviation of three runs, respectively.
\textit{Ours (true+soft)} attains the smallest MSE consistently over the whole prediction horizon due to access to the true graph structures, which serves as a performance upper bound.
\textit{Ours (hybrid)} achieves a smaller MSE than \textit{Ours (full+soft)} / \textit{Ours (ELBO+soft)} and even approaches the performance of \textit{Ours (true+soft)}, which indicates that the RL-based hard attention mechanism indeed helps on the relation recognition, which further improves prediction performance.
\textit{Ours (hybrid)} also outperforms NRI and DNRI where the interaction graph is encoded to a latent space.

\begin{table}[!tbp]
	\centering
	\setlength{\tabcolsep}{1.2mm}{
		\resizebox{\columnwidth}{!}{
			\begin{tabular}{l|cccc|c}
				\hline
				Method & 1.0s & 2.0s &3.0s & 4.0s & MR\\
				\hline 
				NRI \cite{kipf2018neural} & 0.19/0.23 & 0.37/0.57 & 0.58/1.00 & 0.82/1.52 & 27.1\\
				DNRI \cite{graber2020dynamic} & 0.17/0.20 & 0.33/0.49 & 0.50/0.87 & 0.71/1.33 & 22.4\\
				STGAT~\cite{huang2019stgat} & 0.16/0.19 & 0.31/0.47 & 0.48/0.83 & 0.68/1.27 & 20.6 \\
				S-STGCNN~\cite{mohamed2020social} & 0.18/0.21 & 0.34/0.52 & 0.53/0.92 & 0.75/1.41 & 24.7\\
				\hline
				Ours (full+soft) & 0.07/0.10 & 0.17/0.41 & 0.40/0.82 & 0.63/1.28 & 21.0\\
				Ours (ELBO+soft) & \textbf{0.04}/\textbf{0.07} & 0.16/0.42 & 0.40/0.79 & 0.61/1.24 & 16.1\\
				Ours (hybrid, static) & 0.05/0.09 & 0.15/0.40 & 0.38/0.77 & 0.58/1.17 & 11.3\\
				Ours (hybrid, dynamic) & 0.06/0.09 & \textbf{0.14}/\textbf{0.38} & \textbf{0.37}/\textbf{0.75} & \textbf{0.54}/\textbf{1.12} & \textbf{9.5}\\
				\hline
	\end{tabular}}}
	\vspace{0.1cm}
	\caption{Comparison of minADE$_{20}$ / minFDE$_{20}$ (Meters) and Miss Rate@2.0m (MR, \%) of Vehicle Trajectory Prediction.}\label{tab:nuScenes_veh} 
	\vspace{-0.3cm}
\end{table}

We also visualize the predicted trajectories and attention maps of typical cases in Figure \ref{fig:particle_plot} with descriptions in the caption.
The results show that the RL agent can figure out truly relevant agents with a high accuracy and the soft attention counterpart further determines the relative significance of the selected entities.
However, applying soft attention alone on fully connected graphs sometimes assigns trivial weights or even large weights to irrelevant agents as shown in the ``Soft Attention" columns, which weakens interpretability and performance of the model.
We address two interesting cases in (e) and (f), where the RL agent only selects a subset of interacting agents or even an empty set, but the system still generates very good prediction. The reason is that the distance between the interacting particles is too large, which leads to the ignorance of the weak forces by the RL agent. Particularly in (f), all the particles perform nearly uniform motions which results in an empty hybrid attention map.
This demonstrates the capability of distinguishing key information of the hard attention mechanism.

\begin{table}[!tbp]
	\centering
	\setlength{\tabcolsep}{1.2mm}{
		\resizebox{\columnwidth}{!}{
			\begin{tabular}{l|cccc|c}
				\hline
				Method & 1.0s & 2.0s &3.0s &4.0s & MR\\
				\hline 
				NRI \cite{kipf2018neural} & 0.10/0.13 & 0.22/0.32 & 0.35/0.58 & 0.48/0.82 & 28.5\\
				DNRI \cite{graber2020dynamic} & 0.09/0.12 & 0.20/0.31 & 0.32/0.53 & 0.44/0.78 & 25.6\\
				STGAT~\cite{huang2019stgat} & 0.08/0.11 & 0.18/0.27 & 0.29/0.48 & 0.40/0.71 & 17.8 \\
				S-STGCNN~\cite{mohamed2020social} & 0.12/0.15 & 0.23/0.37 & 0.35/0.61 & 0.48/0.88 & 27.9 \\
				\hline
				Ours (full+soft) & 0.07/0.11 & 0.15/0.22 & 0.24/0.40 & 0.32/0.56 & 13.2\\
				Ours (ELBO+soft) & \textbf{0.04}/0.06 & 0.13/0.20 & 0.21/0.39 & 0.30/0.56 & 12.9\\
				Ours (hybrid, static) & 0.06/0.08 & 0.12/0.20 & 0.19/0.36 & 0.29/0.54 & 11.7\\
				Ours (hybrid, dynamic) & 0.05/0.08 & \textbf{0.11}/0.18 & \textbf{0.17}/\textbf{0.34} & \textbf{0.26}/\textbf{0.51} & \textbf{9.8} \\
				\hline
	\end{tabular}}}
	\vspace{0.1cm}
	\caption{Comparison of minADE$_{20}$ / minFDE$_{20}$ (Meters) and Miss Rate@1.0m (MR, \%) of Pedestrian Trajectory Prediction.}
	\vspace{-0.3cm}
	\label{tab:nuScenes_ped}
\end{table}

\begin{figure*}[htbp]
	\centering
	\includegraphics[width=\textwidth]{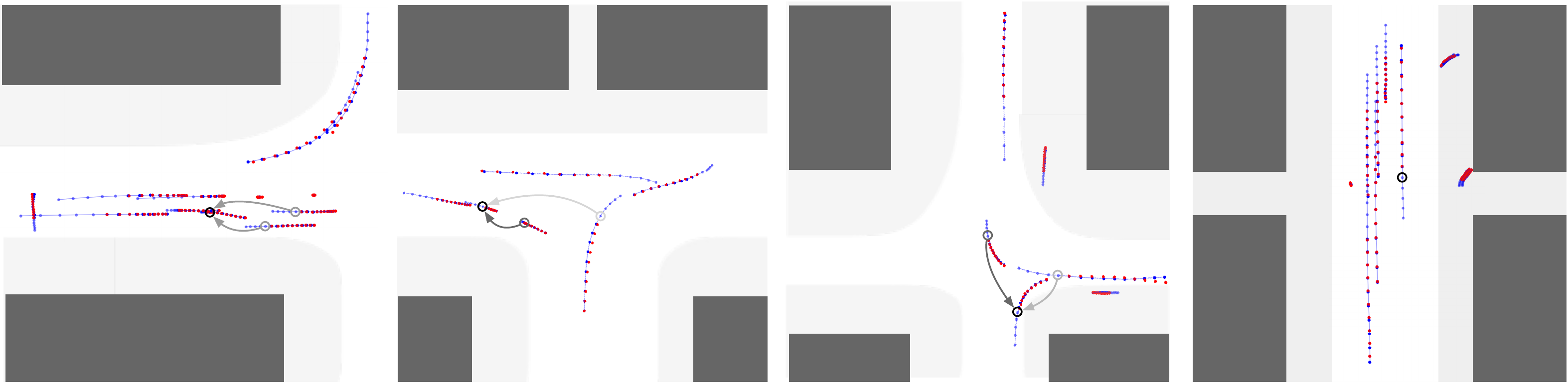}
	\caption{The visualization of testing cases in the nuScenes dataset. The light blue dots are history observations, dark blue dots are ground truth, and red dots are predictions with the minimum ADE. The black circles indicate the target agent and gray arrows indicate hybrid attention. The targets only attend to the agents with arrows selected by hard attention and darker colors imply larger weights. In the last case, there is no arrow, which implies that the model infers that the target is not influenced by any agent in the scene at the current frame.}
	\label{fig:nuScenes_plot}
\end{figure*}

\begin{table*}[htbp]
	\centering
	\setlength{\tabcolsep}{1.5mm}{
		\resizebox{\textwidth}{!}{
			\begin{tabular}{l|cccc|c|cccc|c|cccc|c|cccc|c}
				\hline
				Motion & \multicolumn{5}{c|}{Walking} & \multicolumn{5}{c}{Eating} & \multicolumn{5}{|c|}{Smoking} & \multicolumn{5}{c}{Discussion}\\
				\hline
				milliseconds & 80&160&320&400&1k & 80&160&320&400&1k & 80&160&320&400&1k & 80&160&320&400&1k \\
				\hline
				Res-sup.~\cite{martinez2017human} & 0.27 & 0.46 & 0.67 & 0.75 & 1.03 & 0.23 &  0.37 & 0.59 & 0.73 & 1.08 & 0.32 & 0.59 & 1.01 & 1.10 & 1.50 & 0.30 & 0.67 & 0.98 & 1.06 & 1.69\\
				CSM~\cite{li2018convolutional} & 0.33 & 0.54 & 0.68 & 0.73 & 0.92 & 0.22 &  0.36 & 0.58 & 0.71 & 1.24 & 0.26 & 0.49 & 0.96 & 0.92 & 1.62 & 0.32 & 0.67 & 0.94 & 1.01 & 1.86\\
				Traj-GCN~\cite{liu2019towards} & { 0.18} & {0.32} & { 0.49} & { 0.56} & 0.67 & { 0.17} & 0.31 & 0.52 & 0.62 & 1.12 & 0.22 & 0.41 & 0.84 & 0.79 & 1.57 & { 0.20} & \textbf{0.51} & { 0.79} & { 0.86} & 1.70\\
				DMGNN \cite{li2020dynamic} & { 0.18} & { 0.31} & { 0.49} & {0.58} & 0.75 & { 0.17} & { 0.30} & { 0.49} & { 0.59} & 1.14 & { 0.21} & { 0.39} & { 0.81} & { 0.77} & 1.52 & {0.26} & {0.65} & {0.92} & {0.99} & \textbf{1.45}\\
				LTD-10-10 \cite{mao2019learning} & 0.18 & 0.31 & 0.49 & 0.56 & 0.64 & \textbf{0.16} & 0.29 & 0.50 & 0.62 & 1.10 & 0.22 & 0.41 & 0.86 & 0.80 & 1.58 & 0.20 & \textbf{0.51} & \textbf{0.77} & \textbf{0.85} & 1.75 \\
				HisRepItself \cite{wei2020his} & 0.18 & 0.30 & 0.46 & 0.51 & 0.64 & \textbf{0.16} & 0.29 & 0.49 & 0.60 & 1.10 & 0.22 & 0.42 & 0.86 & 0.80 & 1.58 & 0.20 & 0.52 & 0.78 & 0.87 & 1.63 \\
				\hline
				Ours (hybrid) & \textbf{0.17} & \textbf{0.28} & \textbf{0.43} & \textbf{0.49} & \textbf{0.60} & \textbf{0.16} & \textbf{0.26} & \textbf{0.45} & \textbf{0.56} & \textbf{1.01} & \textbf{0.20} & \textbf{0.38} & \textbf{0.80} & \textbf{0.74} & \textbf{1.48} & \textbf{0.19} & 0.52 & 0.82 & 0.91 & 1.64\\
				\hline
	\end{tabular}}}
	\vspace{0.1cm}
	\caption{Comparison of mean angle errors (MAE) of different methods for both short-term ($\leq$ 400 milliseconds) and long-term prediction (1k milliseconds) on four representative actions in the Human3.6M dataset.}
	\label{tab:pred_h36m_4}
\end{table*}

\subsection{nuScenes Dataset: Traffic Scenarios}\label{sec:nuScenes_main}

We validated our RAIN framework on the nuScenes dataset, which handles long-term prediction of heterogeneous traffic participants (i.e., vehicles and pedestrians).
We predicted the future 4.0s (20 frames) given the history observations of 2.0s (10 frames).
The comparison of quantitative results is shown in Table \ref{tab:nuScenes_veh} (vehicles) and Table \ref{tab:nuScenes_ped} (pedestrians).
Note that we simultaneously included both types of agents during training, while reported the test results separately as done in \cite{salzmann2020trajectron++}.

All the learning-based baseline methods consider the relation modeling via certain techniques. 
It is shown that \textit{Ours (hybrid, dynamic)} achieves the smallest error and lowest miss rate in both vehicle and pedestrian prediction.
We also present extensive ablation results.
\textit{Ours (full+soft)} achieves lower prediction error than STGAT with a similar model structure, especially in vehicle prediction. The reason is that the vehicle kinematics model is incorporated into the motion generator to propagate the vehicle state, which can guarantee the feasibility of generated trajectories.
\textit{Ours (hybrid, static)} and \textit{Ours (ELBO+soft)} outperforms \textit{Ours (full+soft)}, which implies the effectiveness of hard attention on selecting important agents.
In particular, \textit{Ours (hybrid, static)} achieves smaller prediction errors than \textit{Ours (ELBO+soft)}, which implies the advantage of RL based hard attention over the ELBO based method. 
\textit{Ours (hybrid, dynamic)} further reduces the error owing to the dynamic recognition of key elements, where the margin of improvement becomes larger as the prediction horizon increases.

The visualization of agent trajectories and hybrid attention weights in several typical test scenarios is provided in Figure \ref{fig:nuScenes_plot}. 
It is shown that our method can learn reasonable attention weights to exploit the key information, and generate plausible and accurate prediction hypotheses.

\subsection{Human3.6M Dataset: Human Motions}\label{sec:human}

We adopted the same experimental setting as the baseline approaches for fair comparison, which is to forecast the future 25 frames based on the past 50 frames. The model was trained to only predict 10 frames during the training phase, while the prediction was recursively applied by using predictions as new observations during the testing phase.
The forecasting results are shown in Table \ref{tab:pred_h36m_4}, where we only include four representative actions due to space limit. The complete table can be found in Table \ref{tab:pred_h36m_11} in the supplementary materials. 
Among the baselines, HisRepItself \cite{wei2020his} yields the previous state-of-the-art performance, which also serves as an ablation model with only soft attention on the motion history. 
In general, the results show that \textit{Ours (hybrid)} achieves the best performance both in average and for most actions in terms of both short-term and long-term forecasting accuracy. 
In particular, \textit{Ours (hybrid)} outperforms HisRepItself in average, indicating the additional benefits brought by the hard attention mechanism.

\vspace{-0.1cm}
\section{Conclusion}
In this paper, we present a generic motion forecasting framework with key information selection and ranking procedures.
The former is realized by a reinforcement learning based hard attention mechanism for the purpose of discriminating relevant information for the prediction from complete spatio-temporal observations.
The latter is fulfilled by a soft attention mechanism for the purpose of determining relative importance of the selected key elements.
These dual procedures can be applied recurrently to dynamically adjust the focus of the model as the situation evolves over time.
A double-stage training pipeline with an alternating training strategy is employed to train different parts of the model, which proves to be effective and stable in the experiments. 
The general framework is instantiated and applied to multi-agent trajectory prediction and human motion forecasting, which achieves state-of-the-art performance in terms of a wide range of evaluation metrics in different domains.

{\small
	\bibliographystyle{ieee_fullname}
	\bibliography{egbib}

\begin{thebibliography}{10}\itemsep=-1pt

\bibitem{alahi2016social}
Alexandre Alahi, Kratarth Goel, Vignesh Ramanathan, Alexandre Robicquet, Li
  Fei-Fei, and Silvio Savarese.
\newblock Social lstm: Human trajectory prediction in crowded spaces.
\newblock In {\em Proceedings of the IEEE conference on computer vision and
  pattern recognition}, pages 961--971, 2016.

\bibitem{anderson2018bottom}
Peter Anderson, Xiaodong He, Chris Buehler, Damien Teney, Mark Johnson, Stephen
  Gould, and Lei Zhang.
\newblock Bottom-up and top-down attention for image captioning and visual
  question answering.
\newblock In {\em Proceedings of the IEEE conference on computer vision and
  pattern recognition}, pages 6077--6086, 2018.

\bibitem{battaglia2016interaction}
Peter Battaglia, Razvan Pascanu, Matthew Lai, Danilo~Jimenez Rezende, et~al.
\newblock Interaction networks for learning about objects, relations and
  physics.
\newblock In {\em Advances in neural information processing systems}, pages
  4502--4510, 2016.

\bibitem{nuscenes2019}
Holger Caesar, Varun Bankiti, Alex~H. Lang, Sourabh Vora, Venice~Erin Liong,
  Qiang Xu, Anush Krishnan, Yu Pan, Giancarlo Baldan, and Oscar Beijbom.
\newblock nuscenes: A multimodal dataset for autonomous driving.
\newblock {\em arXiv preprint arXiv:1903.11027}, 2019.

\bibitem{chai2020multipath}
Yuning Chai, Benjamin Sapp, Mayank Bansal, and Dragomir Anguelov.
\newblock Multipath: Multiple probabilistic anchor trajectory hypotheses for
  behavior prediction.
\newblock In {\em Conference on Robot Learning}, pages 86--99, 2020.

\bibitem{cho2020self}
Sangwoo Cho, Muhammad Maqbool, Fei Liu, and Hassan Foroosh.
\newblock Self-attention network for skeleton-based human action recognition.
\newblock In {\em The IEEE Winter Conference on Applications of Computer
  Vision}, pages 635--644, 2020.

\bibitem{choi2021shared}
Chiho Choi, Joon~Hee Choi, Jiachen Li, and Srikanth Malla.
\newblock Shared cross-modal trajectory prediction for autonomous driving.
\newblock In {\em Proceedings of the IEEE/CVF Conference on Computer Vision and
  Pattern Recognition}, pages 244--253, 2021.

\bibitem{choi2019looking}
Chiho Choi and Behzad Dariush.
\newblock Looking to relations for future trajectory forecast.
\newblock In {\em Proceedings of the IEEE International Conference on Computer
  Vision}, pages 921--930, 2019.

\bibitem{deo2018convolutional}
Nachiket Deo and Mohan~M Trivedi.
\newblock Convolutional social pooling for vehicle trajectory prediction.
\newblock In {\em Proceedings of the IEEE Conference on Computer Vision and
  Pattern Recognition Workshops}, pages 1468--1476, 2018.

\bibitem{fan2019multi}
Chenyou Fan, Yuze Zhang, Yi Pan, Xiaoyue Li, Chi Zhang, Rong Yuan, Di Wu,
  Wensheng Wang, Jian Pei, and Heng Huang.
\newblock Multi-horizon time series forecasting with temporal attention
  learning.
\newblock In {\em Proceedings of the 25th ACM SIGKDD International Conference
  on Knowledge Discovery \& Data Mining}, pages 2527--2535, 2019.

\bibitem{fragkiadaki2015recurrent}
Katerina Fragkiadaki, Sergey Levine, Panna Felsen, and Jitendra Malik.
\newblock Recurrent network models for human dynamics.
\newblock In {\em Proceedings of the IEEE International Conference on Computer
  Vision}, pages 4346--4354, 2015.

\bibitem{pmlr-v97-franceschi19a}
Luca Franceschi, Mathias Niepert, Massimiliano Pontil, and Xiao He.
\newblock Learning discrete structures for graph neural networks.
\newblock In {\em Proceedings of the 36th International Conference on Machine
  Learning}, pages 1972--1982, 2019.

\bibitem{gang2021loki}
Haiming Gang, Harshayu Girase, Srikanth Malla, Jiachen Li, Akira Kanehara,
  Karttikeya Mangalam, and Chiho Choi.
\newblock Loki: Long term and key intentions for trajectory prediction.
\newblock In {\em Proceedings of the International Conference on Computer
  Vision (ICCV)}, 2021.

\bibitem{gao2019graph}
Hongyang Gao and Shuiwang Ji.
\newblock Graph representation learning via hard and channel-wise attention
  networks.
\newblock In {\em Proceedings of the 25th ACM SIGKDD International Conference
  on Knowledge Discovery \& Data Mining}, pages 741--749, 2019.

\bibitem{gao2020vectornet}
Jiyang Gao, Chen Sun, Hang Zhao, Yi Shen, Dragomir Anguelov, Congcong Li, and
  Cordelia Schmid.
\newblock Vectornet: Encoding hd maps and agent dynamics from vectorized
  representation.
\newblock In {\em Proceedings of the IEEE/CVF Conference on Computer Vision and
  Pattern Recognition}, pages 11525--11533, 2020.

\bibitem{gopalakrishnan2019neural}
Anand Gopalakrishnan, Ankur Mali, Dan Kifer, Lee Giles, and Alexander~G
  Ororbia.
\newblock A neural temporal model for human motion prediction.
\newblock In {\em Proceedings of the IEEE Conference on Computer Vision and
  Pattern Recognition}, pages 12116--12125, 2019.

\bibitem{graber2020dynamic}
Colin Graber and Alexander~G Schwing.
\newblock Dynamic neural relational inference.
\newblock In {\em Proceedings of the IEEE/CVF Conference on Computer Vision and
  Pattern Recognition}, pages 8513--8522, 2020.

\bibitem{hasselt2016deep}
Hado~van Hasselt, Arthur Guez, and David Silver.
\newblock Deep reinforcement learning with double q-learning.
\newblock In {\em Proceedings of the Thirtieth AAAI Conference on Artificial
  Intelligence}, pages 2094--2100. AAAI Press, 2016.

\bibitem{hoshen2017vain}
Yedid Hoshen.
\newblock Vain: Attentional multi-agent predictive modeling.
\newblock In {\em Advances in Neural Information Processing Systems}, pages
  2701--2711, 2017.

\bibitem{huang2019stgat}
Yingfan Huang, Huikun Bi, Zhaoxin Li, Tianlu Mao, and Zhaoqi Wang.
\newblock Stgat: Modeling spatial-temporal interactions for human trajectory
  prediction.
\newblock In {\em Proceedings of the IEEE International Conference on Computer
  Vision}, pages 6272--6281, 2019.

\bibitem{h36m_pami}
Catalin Ionescu, Dragos Papava, Vlad Olaru, and Cristian Sminchisescu.
\newblock Human3.6m: Large scale datasets and predictive methods for 3d human
  sensing in natural environments.
\newblock {\em IEEE Transactions on Pattern Analysis and Machine Intelligence},
  36(7):1325--1339, jul 2014.

\bibitem{jain2016structural}
Ashesh Jain, Amir~R Zamir, Silvio Savarese, and Ashutosh Saxena.
\newblock Structural-rnn: Deep learning on spatio-temporal graphs.
\newblock In {\em Proceedings of the ieee conference on computer vision and
  pattern recognition}, pages 5308--5317, 2016.

\bibitem{kipf2018neural}
Thomas Kipf, Ethan Fetaya, Kuan-Chieh Wang, Max Welling, and Richard Zemel.
\newblock Neural relational inference for interacting systems.
\newblock In {\em International Conference on Machine Learning}, pages
  2688--2697, 2018.

\bibitem{kosaraju2019social}
Vineet Kosaraju, Amir Sadeghian, Roberto Mart{\'\i}n-Mart{\'\i}n, Ian Reid,
  Hamid Rezatofighi, and Silvio Savarese.
\newblock Social-bigat: Multimodal trajectory forecasting using bicycle-gan and
  graph attention networks.
\newblock In {\em Advances in Neural Information Processing Systems}, pages
  137--146, 2019.

\bibitem{Kossen2020Structured}
Jannik Kossen, Karl Stelzner, Marcel Hussing, Claas Voelcker, and Kristian
  Kersting.
\newblock Structured object-aware physics prediction for video modeling and
  planning.
\newblock In {\em International Conference on Learning Representations}, 2020.

\bibitem{lee2017desire}
Namhoon Lee, Wongun Choi, Paul Vernaza, Christopher~B Choy, Philip~HS Torr, and
  Manmohan Chandraker.
\newblock Desire: Distant future prediction in dynamic scenes with interacting
  agents.
\newblock In {\em Proceedings of the IEEE Conference on Computer Vision and
  Pattern Recognition}, pages 336--345, 2017.

\bibitem{li2018convolutional}
Chen Li, Zhen Zhang, Wee Sun~Lee, and Gim Hee~Lee.
\newblock Convolutional sequence to sequence model for human dynamics.
\newblock In {\em Proceedings of the IEEE Conference on Computer Vision and
  Pattern Recognition}, pages 5226--5234, 2018.

\bibitem{li2019conditional}
Jiachen Li, Hengbo Ma, and Masayoshi Tomizuka.
\newblock Conditional generative neural system for probabilistic trajectory
  prediction.
\newblock In {\em 2019 IEEE/RSJ International Conference on Intelligent Robots
  and Systems (IROS)}, pages 6150--6156. IEEE, 2019.

\bibitem{li2021spatio}
Jiachen Li, Hengbo Ma, Zhihao Zhang, Jinning Li, and Masayoshi Tomizuka.
\newblock Spatio-temporal graph dual-attention network for multi-agent
  prediction and tracking.
\newblock {\em IEEE Transactions on Intelligent Transportation Systems}, 2021.

\bibitem{li2020evolvegraph}
Jiachen Li, Fan Yang, Masayoshi Tomizuka, and Chiho Choi.
\newblock Evolvegraph: Multi-agent trajectory prediction with dynamic
  relational reasoning.
\newblock In {\em Advances in neural information processing systems}, 2020.

\bibitem{li2020dynamic}
Maosen Li, Siheng Chen, Yangheng Zhao, Ya Zhang, Yanfeng Wang, and Qi Tian.
\newblock Dynamic multiscale graph neural networks for 3d skeleton based human
  motion prediction.
\newblock In {\em Proceedings of the IEEE/CVF Conference on Computer Vision and
  Pattern Recognition}, pages 214--223, 2020.

\bibitem{li2020causal}
Yunzhu Li, Antonio Torralba, Anima Anandkumar, Dieter Fox, and Animesh Garg.
\newblock Causal discovery in physical systems from videos.
\newblock {\em Advances in Neural Information Processing Systems}, 33, 2020.

\bibitem{liu2018picanet}
Nian Liu, Junwei Han, and Ming-Hsuan Yang.
\newblock Picanet: Learning pixel-wise contextual attention for saliency
  detection.
\newblock In {\em Proceedings of the IEEE Conference on Computer Vision and
  Pattern Recognition}, pages 3089--3098, 2018.

\bibitem{liu2019towards}
Zhenguang Liu, Shuang Wu, Shuyuan Jin, Qi Liu, Shijian Lu, Roger Zimmermann,
  and Li Cheng.
\newblock Towards natural and accurate future motion prediction of humans and
  animals.
\newblock In {\em Proceedings of the IEEE Conference on Computer Vision and
  Pattern Recognition}, pages 10004--10012, 2019.

\bibitem{ma2021multi}
Hengbo Ma, Yaofeng Sun, Jiachen Li, and Masayoshi Tomizuka.
\newblock Multi-agent driving behavior prediction across different scenarios
  with self-supervised domain knowledge.
\newblock In {\em 2021 International Conference on Intelligent Transportation
  Systems (ITSC)}. IEEE, 2021.

\bibitem{ma2021continual}
Hengbo Ma, Yaofeng Sun, Jiachen Li, Masayoshi Tomizuka, and Chiho Choi.
\newblock Continual multi-agent interaction behavior prediction with
  conditional generative memory.
\newblock {\em IEEE Robotics and Automation Letters}, 2021.

\bibitem{ma2020reinforcement}
Xiaobai Ma, Jiachen Li, Mykel~J Kochenderfer, David Isele, and Kikuo Fujimura.
\newblock Reinforcement learning for autonomous driving with latent state
  inference and spatial-temporal relationships.
\newblock In {\em 2021 International Conference on Robotics and Automation
  (ICRA)}. IEEE, 2021.

\bibitem{maddison2016concrete}
Chris~J Maddison, Andriy Mnih, and Yee~Whye Teh.
\newblock The concrete distribution: A continuous relaxation of discrete random
  variables.
\newblock In {\em International Conference on Learning Representations}, 2017.

\bibitem{mao2019learning}
Wei Mao, Miaomiao Liu, Mathieu Salzmann, and Hongdong Li.
\newblock Learning trajectory dependencies for human motion prediction.
\newblock In {\em Proceedings of the IEEE International Conference on Computer
  Vision}, pages 9489--9497, 2019.

\bibitem{martinez2017human}
Julieta Martinez, Michael~J Black, and Javier Romero.
\newblock On human motion prediction using recurrent neural networks.
\newblock In {\em Proceedings of the IEEE Conference on Computer Vision and
  Pattern Recognition}, pages 2891--2900, 2017.

\bibitem{mnih2015human}
Volodymyr Mnih, Koray Kavukcuoglu, David Silver, Andrei~A Rusu, Joel Veness,
  Marc~G Bellemare, Alex Graves, Martin Riedmiller, Andreas~K Fidjeland, Georg
  Ostrovski, et~al.
\newblock Human-level control through deep reinforcement learning.
\newblock {\em nature}, 518(7540):529--533, 2015.

\bibitem{mohamed2020social}
Abduallah Mohamed, Kun Qian, Mohamed Elhoseiny, and Christian Claudel.
\newblock Social-stgcnn: A social spatio-temporal graph convolutional neural
  network for human trajectory prediction.
\newblock In {\em Proceedings of the IEEE/CVF Conference on Computer Vision and
  Pattern Recognition}, pages 14424--14432, 2020.

\bibitem{parikh2016decomposable}
Ankur Parikh, Oscar T{\"a}ckstr{\"o}m, Dipanjan Das, and Jakob Uszkoreit.
\newblock A decomposable attention model for natural language inference.
\newblock In {\em Proceedings of the 2016 Conference on Empirical Methods in
  Natural Language Processing}, pages 2249--2255, 2016.

\bibitem{phan2020covernet}
Tung Phan-Minh, Elena~Corina Grigore, Freddy~A Boulton, Oscar Beijbom, and
  Eric~M Wolff.
\newblock Covernet: Multimodal behavior prediction using trajectory sets.
\newblock In {\em Proceedings of the IEEE/CVF Conference on Computer Vision and
  Pattern Recognition}, pages 14074--14083, 2020.

\bibitem{qin2017dual}
Yao Qin, Dongjin Song, Haifeng Cheng, Wei Cheng, Guofei Jiang, and Garrison~W
  Cottrell.
\newblock A dual-stage attention-based recurrent neural network for time series
  prediction.
\newblock In {\em Proceedings of the 26th International Joint Conference on
  Artificial Intelligence}, pages 2627--2633, 2017.

\bibitem{salzmann2020trajectron++}
Tim Salzmann, Boris Ivanovic, Punarjay Chakravarty, and Marco Pavone.
\newblock Trajectron++: Multi-agent generative trajectory forecasting with
  heterogeneous data for control.
\newblock In {\em Proceedings of Europe Conference on Computer Vision (ECCV)},
  2020.

\bibitem{shen2018reinforced}
Tao Shen, Tianyi Zhou, Guodong Long, Jing Jiang, Sen Wang, and Chengqi Zhang.
\newblock Reinforced self-attention network: a hybrid of hard and soft
  attention for sequence modeling.
\newblock In {\em Proceedings of the 27th International Joint Conference on
  Artificial Intelligence}, pages 4345--4352, 2018.

\bibitem{si2019attention}
Chenyang Si, Wentao Chen, Wei Wang, Liang Wang, and Tieniu Tan.
\newblock An attention enhanced graph convolutional lstm network for
  skeleton-based action recognition.
\newblock In {\em Proceedings of the IEEE conference on computer vision and
  pattern recognition}, pages 1227--1236, 2019.

\bibitem{song2018spatio}
Sijie Song, Cuiling Lan, Junliang Xing, Wenjun Zeng, and Jiaying Liu.
\newblock Spatio-temporal attention-based lstm networks for 3d action
  recognition and detection.
\newblock {\em IEEE Transactions on image processing}, 27(7):3459--3471, 2018.

\bibitem{tang2019multiple}
Charlie Tang and Russ~R Salakhutdinov.
\newblock Multiple futures prediction.
\newblock In {\em Advances in Neural Information Processing Systems}, pages
  15424--15434, 2019.

\bibitem{tang2018long}
Yongyi Tang, Lin Ma, Wei Liu, and Wei-Shi Zheng.
\newblock Long-term human motion prediction by modeling motion context and
  enhancing motion dynamic.
\newblock In {\em Proceedings of the 27th International Joint Conference on
  Artificial Intelligence}, pages 935--941, 2018.

\bibitem{van2018relational}
Sjoerd van Steenkiste, Michael Chang, Klaus Greff, and J{\"u}rgen Schmidhuber.
\newblock Relational neural expectation maximization: Unsupervised discovery of
  objects and their interactions.
\newblock In {\em International Conference on Learning Representations}, 2018.

\bibitem{vel2018graph}
Petar Veličković, Guillem Cucurull, Arantxa Casanova, Adriana Romero, Pietro
  Liò, and Yoshua Bengio.
\newblock Graph attention networks.
\newblock In {\em International Conference on Learning Representations}, 2018.

\bibitem{vemula2018social}
Anirudh Vemula, Katharina Muelling, and Jean Oh.
\newblock Social attention: Modeling attention in human crowds.
\newblock In {\em 2018 IEEE international Conference on Robotics and Automation
  (ICRA)}, pages 1--7. IEEE, 2018.

\bibitem{wang2020active}
Boyu Wang, Lihan Huang, and Minh Hoai.
\newblock Active vision for early recognition of human actions.
\newblock In {\em Proceedings of the IEEE/CVF Conference on Computer Vision and
  Pattern Recognition}, pages 1081--1091, 2020.

\bibitem{wang2017residual}
Fei Wang, Mengqing Jiang, Chen Qian, Shuo Yang, Cheng Li, Honggang Zhang,
  Xiaogang Wang, and Xiaoou Tang.
\newblock Residual attention network for image classification.
\newblock In {\em Proceedings of the IEEE conference on computer vision and
  pattern recognition}, pages 3156--3164, 2017.

\bibitem{wang2016attention}
Yequan Wang, Minlie Huang, Xiaoyan Zhu, and Li Zhao.
\newblock Attention-based lstm for aspect-level sentiment classification.
\newblock In {\em Proceedings of the 2016 conference on empirical methods in
  natural language processing}, pages 606--615, 2016.

\bibitem{watters2017visual}
Nicholas Watters, Daniel Zoran, Theophane Weber, Peter Battaglia, Razvan
  Pascanu, and Andrea Tacchetti.
\newblock Visual interaction networks: Learning a physics simulator from video.
\newblock In {\em Advances in neural information processing systems}, pages
  4539--4547, 2017.

\bibitem{wei2020his}
Mao Wei, Liu Miaomiao, and Salzemann Mathieu.
\newblock History repeats itself: Human motion prediction via motion attention.
\newblock In {\em ECCV}, 2020.

\bibitem{williams1992simple}
Ronald~J Williams.
\newblock Simple statistical gradient-following algorithms for connectionist
  reinforcement learning.
\newblock {\em Machine learning}, 8(3-4):229--256, 1992.

\bibitem{wu2020motionnet}
Pengxiang Wu, Siheng Chen, and Dimitris~N Metaxas.
\newblock Motionnet: Joint perception and motion prediction for autonomous
  driving based on bird's eye view maps.
\newblock In {\em Proceedings of the IEEE/CVF Conference on Computer Vision and
  Pattern Recognition}, pages 11385--11395, 2020.

\bibitem{wu2019multi}
Wenhao Wu, Dongliang He, Xiao Tan, Shifeng Chen, and Shilei Wen.
\newblock Multi-agent reinforcement learning based frame sampling for effective
  untrimmed video recognition.
\newblock In {\em Proceedings of the IEEE International Conference on Computer
  Vision}, pages 6222--6231, 2019.

\bibitem{xu2015show}
Kelvin Xu, Jimmy Ba, Ryan Kiros, Kyunghyun Cho, Aaron Courville, Ruslan
  Salakhudinov, Rich Zemel, and Yoshua Bengio.
\newblock Show, attend and tell: Neural image caption generation with visual
  attention.
\newblock In {\em International conference on machine learning}, pages
  2048--2057, 2015.

\bibitem{zhao2019pyramid}
Ting Zhao and Xiangqian Wu.
\newblock Pyramid feature attention network for saliency detection.
\newblock In {\em Proceedings of the IEEE Conference on Computer Vision and
  Pattern Recognition}, pages 3085--3094, 2019.

\end{thebibliography}
}

\clearpage

\section{Double-Stage Training Pipeline}\label{sec:training_alg}

In this section, we elaborate the training procedures of RAIN for multi-agent interacting systems, which consist of two stages: pre-training stage and formal-training stage.

\subsection{Pre-Training Stage}

\subsubsection{GMP Module}

\begin{figure}[htbp]
	\centering
	\includegraphics[width=\columnwidth]{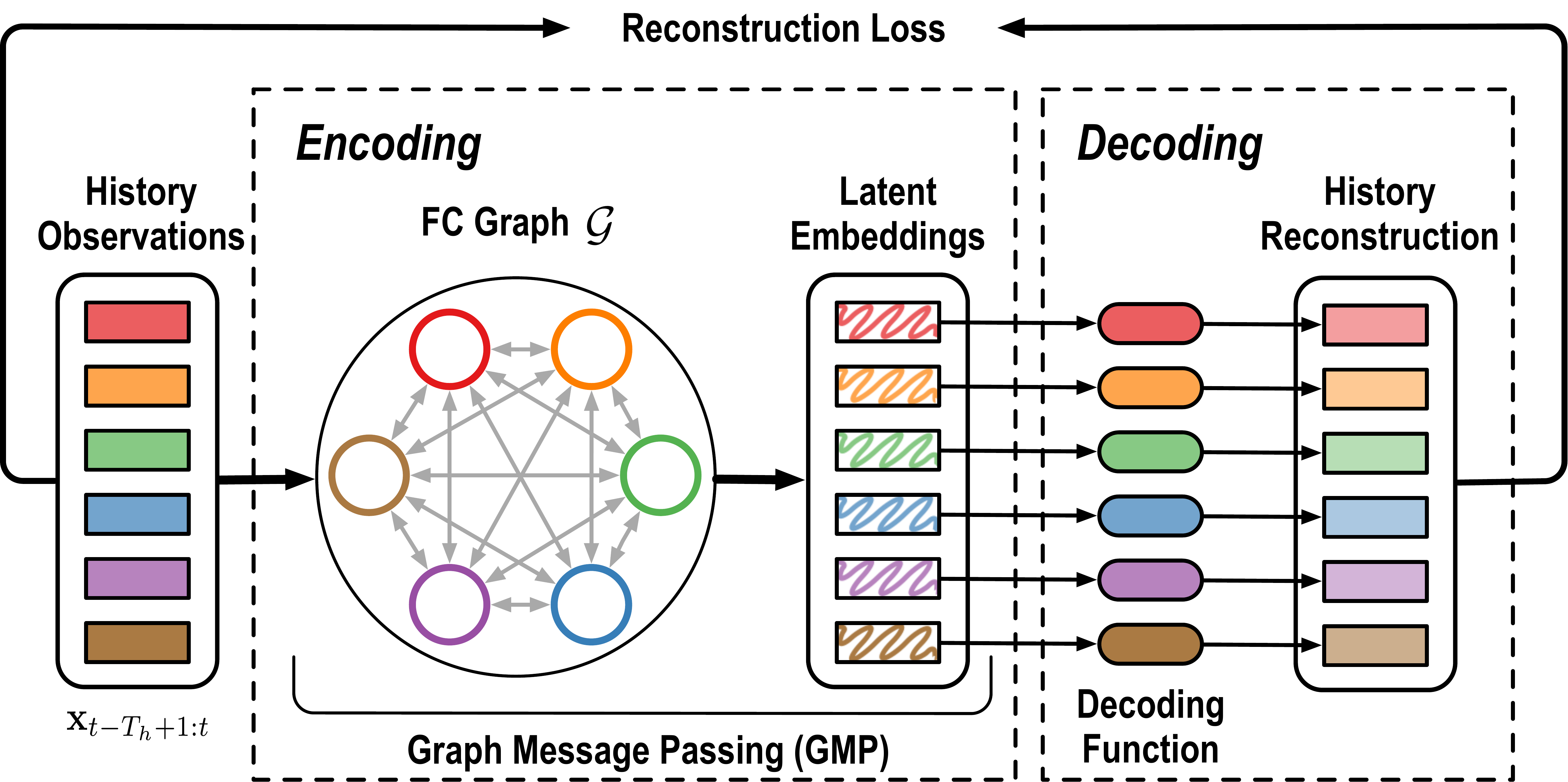}
	\caption{The diagram of the auto-encoder structure for pre-training the GMP module, which consists of an encoding procedure and a decoding procedure.}
	\label{fig:GMP_AE}
\end{figure}

\begin{algorithm}[htbp]\label{alg:training}
	\caption{Double-Stage Training Algorithm}
	\KwIn{history $\mathbf{X}_{t-T_h+1:t}$, true future $\mathbf{X}_{t+1:t+T_f}$, context $\mathbf{C}$, hyperparameters $N_{ft}$, $N_{s}$, $E$}
	Initialize the parameters in GMP ($\phi$), RL-HA ($\psi$) and SGA-MG ($\theta$)\;
	\tcc{Pre-training Stage}
	Pre-train GMP by unsupervised learning with (\ref{equ:GMP_loss})\;
	Pre-train SGA-MG by supervised learning with (\ref{equ:MG_loss})\;
	\tcc{Initialize RL}
	Initialize the replay buffer $\mathcal{D}$\; 
	Initialize the RL-step index $i \leftarrow 0$\;
	\tcc{Formal-training Stage}
	\For{epoch $\leftarrow$ 1,2,...,$E$}{
		Generate rollout $\xi$ with $\phi,\psi,\theta$\;
		Add rollout $\xi$ into replay buffer $\mathcal{D}$\;
		$i \leftarrow i + 1$\;
		\tcc{Train RL-HA}
		\If{$i$ $>$ $N_\text{s}$}{
			Sample a rollout $\xi'$ from $\mathcal{D}$\;
			Update policy and $\psi$ with DDQN\;
		}
		\tcc{Finetune SGA-MG}
		\For{$m \leftarrow 1,2,...,N_\text{ft}$}
		{
			Sample a case of $\mathbf{X}_{t-T_h+1:t}$ and $\mathbf{C}$\;
			Use GMP to obtain node attributes on $\mathcal{G}$\;
			Use RL-HA to generate $\mathcal{G}'$\;
			Use SGA-MG to generate $\hat{\mathbf{X}}_{t+1:t+T_f}$\;
			Compute loss by equation (\ref{equ:MG_loss})\;
			Update $\theta$ by back-propagation\;
		}
	}
	\vspace{-0.cm}
	
\end{algorithm}

We employ a standard encoder-decoder structure to pre-train the GMP by unsupervised learning, with the purpose of enabling informative and effective feature extraction in the GMP module. 
In the auto-encoder structure, the GMP module serves as the encoding process to generate latent embeddings for each node. And an auxiliary decoding function is trained to reconstruct the history information with the latent embeddings. After the model is well-trained, the GMP module is able to extract good representation of the observation information.

The loss function is the standard mean squared error reconstruction loss, which is calculated as
\begin{equation}\label{equ:GMP_loss}
	L_{\text{GMP}} = \frac{1}{NT_h} \sum_{i=1}^{N} \sum_{t'=t-T_h+1}^{t} ||\mathbf{x}^i_{t'}-\hat{\mathbf{x}}^i_{t'}||^2.
\end{equation}
After convergence, we save the parameters of the GMP module and discard those of the decoder, since we only use GMP in the following formal-training stage.

\subsubsection{SGA-MG Module}

In order to enable informative initial reward for the RL-HA module, we pre-train the SGA-MG module with a fully connected topology. The model architecture is the same as in Figure \ref{fig:model_diagram} except that the GAT is applied to a fully connected graph. The loss function is a standard mean squared error loss, which is calculated as
\begin{equation}\label{equ:MG_loss}
	L_{\text{SGA-MG}} = \frac{1}{NT_f} \sum_{i=1}^{N} \sum_{t'=t+1}^{t+T_f} ||\mathbf{x}^i_{t'}-\hat{\mathbf{x}}^i_{t'}||^2.
\end{equation}

\subsection{Formal-Training Stage}

In the formal-training stage, we initialize the GMP and SGA-MG with pre-trained parameters. Then we perform an alternating training strategy to train the RL-HA and SGA-MG alternatively until convergence.
The detailed pseudo-code of the training pipeline of the whole framework is provided in Algorithm \ref{alg:training}.

\section{RAIN for Human Skeleton Motions (Cont.)}\label{sec:RAIN_human}

For human motion forecasting, we employ the state-of-the-art model \cite{wei2020his} as the soft attention based motion generator in our framework. 
We strongly encourage the readers to refer to \cite{wei2020his} for better understanding the model details below.
Similar to RAIN for the multi-agent interacting systems, we also employ a double-stage training pipeline, including a pre-training stage and a formal-training stage. In the pre-training stage, we pre-train the parameters of a contextual encoder and the soft attention based motion generator. In the formal-training stage, we train the RL-HA module and finetune the motion generator alternatively.

We denote the complete history motions as $\mathbf{X}_{t-T_h+1:t}$ and the future motions as $\mathbf{X}_{t+1:t+T_f}$. We have the same assumption as \cite{wei2020his} that $T_h > T_s+T_f$ where $T_s$ is the length of the motion segments used to compute attention weights.

\subsection{Pre-Training Stage}
First, we use an auto-encoder structure to train an encoding function that can extract the contextual information from the complete history motion sequence. More formally, the auto-encoder can be written as
\begin{align}
	\mathbf{Z} = \text{Encoding}(\mathbf{X}_{t-T_h+1:t}),\\
	\hat{\mathbf{X}}_{t-T_h+1:t} = \text{Decoding}(\mathbf{Z}),
\end{align}
where Encoding and Decoding functions are neural networks.
The loss function of training the auto-encoder is the standard mean squared error reconstruction loss, which is calculated by
\begin{align}
	\text{MSE} = \frac{1}{JT_h} \sum_{j=1}^{J}\sum_{t'=t-T_h+1}^{t}||\mathbf{x}^j_{t'} - \hat{\mathbf{x}}^j_{t'}||^2,
\end{align}
where $J$ is the number of relative angles between joints in a skeleton.
For the soft attention based motion generator, since the authors of \cite{wei2020his} released their official code and pre-trained models, we directly load their pre-trained parameters in the formal-training stage.

\subsection{Formal-Training Stage}

In the formal-training stage, we alternatively train the RL-HA module and the motion generator.
We define the motion segments in the same way as \cite{wei2020his}. 
More specifically, we first divide the complete motion history $\mathbf{X}_{t-T_h+1:t}$ into $T_h-T_s-T_f+1$ segments $\{\mathbf{X}_{i:i+T_s+T_f-1}\}^{t-T_s-T_f+1}_{i=t-T_h+1}$, each of which contains $T_s+T_f$ consecutive frames of human poses.
We use the same setting as \cite{wei2020his}, where the motion generator exploits the past $T_s$ frames to predict the future $T_f$ frames.
The first $T_s$ frames of each segment is used as a key, and the whole segment is then the corresponding value. The query is defined as the latest segment $\mathbf{X}_{t-T_s+1:t}$.

\subsubsection{RL-HA Module}
In the domain of forecasting human skeleton motions, the RL-HA module is expected to select the key history motion segments for the current prediction based on the latest observation segment. Then the soft attention mechanism in \cite{wei2020his} will further rank the relative importance of the selected key segments, which is employed by the motion predictor to generate future motions.

The selection of key segments naturally fits into a reinforcement learning framework.
The definition of observations, actions and reward functions of the RL-agent are elaborated in the following.

\textbf{Observations}:
The observation $O$ of RL-agent at RL-step $\eta \ (\leq T_{\text{RL}})$ includes a tuple of key, query, contextual information $\mathbf{Z}$ as well as the current segment selection status $s_{i}$ (0: ``retained" or 1: ``discarded"). $T_{\text{RL}}$ is the upper bound of RL-steps.
The observation $O_\eta$ is obtained by 
\begin{align}
	O_\eta = \left[f_k(\mathbf{X}_{i:i+T_s-1}),\ f_q(\mathbf{X}_{t-T_s+1:t}),\ \mathbf{Z},\ s_{i,\eta}\right],
\end{align}
where $f_k$ and $f_q$ are mapping functions modeled by neural networks.
Note that the dimension of $O_\eta$ only depends on the dimensions of key, query and contextual information, which enables the applicability to the scenarios with varying numbers of history motion segments.
The policy network of RL-agent takes the observation $O_{\eta}$ as input and decides the action at each RL-step.

\textbf{Actions}:
There are two possible actions for the RL-agent: ``staying the same" (action 0) and ``changing to the opposite" (action 1).
At each RL-step, the RL-agent makes decision for each history motion segment. 
The policy can be written as $a=\pi(O)$.
We do not enforce any constraints on the selection of motion segments, i.e., there is no lower / upper bound on the number of selected segments.
The actions of RL-agent may change the key motion segments after each RL-step, which further influences the soft attention based motion generator.

\textbf{Rewards}:
The reward consists of two parts: regular reward $R_\text{reg}$ and improvement reward $R_\text{imp}$.
More specifically, the \textit{regular reward} is the negative mean squared error of future predictions calculated by
\begin{align}
	R_{\text{reg},\eta} = - \frac{1}{J}  \sum_{j=1}^{J} \sum_{t'=t+1}^{t+T_f}||\mathbf{x}^j_{t'} - \hat{\mathbf{x}}^j_{t',\eta}||^2.
\end{align}
The \textit{improvement reward} encourages the decrease of prediction error via applying a sign function to the error change between consecutive RL-steps, which is obtained by
\begin{align}
	R_{\text{imp},\eta} =  \mathrm{sign}(R_{\text{reg},\eta} - R_{\text{reg},\eta-1}).
\end{align}
The whole reward is obtained by $R_{\eta} = R_{\text{reg},\eta} + \beta_\text{imp} R_{\text{imp},\eta}$, where $\beta_\text{imp}$ is a hyperparameter.

\begin{table*}[!tbp]
	\centering
	\setlength{\tabcolsep}{1.5mm}{
		\resizebox{\textwidth}{!}{
			\begin{tabular}{l|cccc|c|cccc|c|cccc|c|cccc|c}
				\hline
				Motion & \multicolumn{5}{c}{Directions} & \multicolumn{5}{|c|}{Greeting} & \multicolumn{5}{c}{Phoning} 
				& \multicolumn{5}{|c}{Posing} \\
				\hline
				millisecond & 80&160&320&400&1k & 80&160&320&400&1k & 80&160&320&400&1k & 80&160&320&400&1k  \\
				\hline
				Res-sup~\cite{martinez2017human} & 0.41 & 0.64 & 0.80 & 0.92 &-- &0.57 & 0.83 & 1.45 & 1.60 &--& 0.59 & 1.06 & 1.45 & 1.60 &--& 0.45 & 0.85 & 1.34 & 1.56 & -- \\
				CSM~\cite{li2018convolutional} & 0.39 & 0.60 & 0.80 & 0.91 &1.45& 0.51 & 0.82 & 1.21 & 1.38 &1.72& 0.59 & 1.13 & 1.51 & 1.65 &1.81& 0.29 & 0.60 & 1.12 & 1.37 & 2.65 \\
				Traj-GCN~\cite{liu2019towards} & {0.26} & {0.45} & 0.70 & 0.79 &-- &{\bf 0.35} & { 0.61} & {0.96} & {1.13} & --&{ 0.53} & {1.02} & 1.32 & 1.45 &--& 0.23 & 0.54 & 1.26 & 1.38 & -- \\
				DMGNN \cite{li2020dynamic} & {\bf 0.25} & { 0.44} & { 0.65} & { 0.71} &--& {0.36} & { 0.61} & {\bf 0.94} & {\bf 1.12} & --&{ 0.52} & { 0.97} & { 1.29} & { 1.43}&-- & { 0.20} & { 0.46} & { 1.06} & { 1.34} &-- \\
				LTD-10-10 \cite{mao2019learning} & 0.26 & 0.45 & 0.71 & 0.79 & 1.35 & 0.36 & \textbf{0.60} & 0.95 & 1.13 & 1.59 & 0.53 & 1.02 & 1.35 & 1.48 & 1.74 & \textbf{0.19} & 0.44 & \textbf{1.01} & 1.24 & 2.55  \\
				HisRepItself \cite{wei2020his} & \textbf{0.25} & \textbf{0.43} & \textbf{0.60} & \textbf{0.69} & \textbf{1.27} & \textbf{0.35} & \textbf{0.60} & 0.95 & 1.14 & \textbf{1.57} & 0.53 & 1.01 & 1.31 & 1.43 & 1.68 & \textbf{0.19} & 0.46 & 1.09 & 1.35 & \textbf{2.32}  \\
				\hline
				Ours (hybrid) &  \textbf{0.25} & 0.46 & 0.65 & 0.75 & 1.34 & \textbf{0.35} & 0.62 & 0.99 & 1.21 & 1.64 & \textbf{0.49} & \textbf{0.95} & \textbf{1.23} & \textbf{1.35} & \textbf{1.59} & \textbf{0.19} & \textbf{0.43} & 1.02 & \textbf{1.23} &2.46 \\
				\hline
				\hline
				Motion & \multicolumn{5}{c|}{Purchases} & \multicolumn{5}{c|}{Sitting} & \multicolumn{5}{c|}{Sitting Down} & \multicolumn{5}{c}{Taking Photo} \\
				\hline
				millisecond & 80&160&320&400&1k & 80&160&320&400&1k & 80&160&320&400&1k & 80&160&320&400&1k  \\
				\hline
				Res-sup~\cite{martinez2017human} & 0.58 & 0.79 & 1.08 & 1.15 &--& 0.41 & 0.68 & 1.12 & 1.33 &--& 0.47 & 0.88 & 1.37 & 1.54 &--& 0.28 & 0.57 & 0.90 & 1.02 &--\\
				CSM~\cite{li2018convolutional} & 0.63 & 0.91 & 1.19 & 1.29 &2.52& 0.39 & 0.61 & 1.02 & 1.18 &1.67& 0.41 & 0.78 & 1.16 & 1.31&2.06 & 0.23 & 0.49 & 0.88 & 1.06 & 1.40\\
				Traj-GCN~\cite{liu2019towards} & {0.42} & 0.66 & 1.04 & 1.12 &-- &0.29 & 0.45 & 0.82 & {0.97} &--& {0.30} & {0.63} & {0.89} & {1.01}&-- & {0.15} & {0.36} & {0.59} & {0.72} &--\\
				DMGNN \cite{li2020dynamic}& {0.41} & {\bf 0.61} & {1.05} & {1.14}&-- & {0.26} & {0.42} & {0.76} & {0.97} &--& {0.32} & {0.65} & {0.93} & {1.05} &-- &{0.15} & {0.34} & {0.58} & {0.71} & --\\
				LTD-10-10 \cite{mao2019learning} & 0.43 & 0.65 & 1.05 & 1.13 & 2.27 & 0.29 & 0.45 & 0.80 & 0.97 & 1.52 & 0.30 & 0.61 & 0.90 & 1.00 & 1.67 & 0.14 & 0.34 & 0.58 & 0.70 & 1.05 \\
				HisRepItself \cite{wei2020his} & 0.42 & 0.65 & \textbf{1.00} & \textbf{1.07} & \textbf{2.22} & 0.29 & 0.47 & 0.83 & 1.01 & 1.55 & 0.30 & 0.63 & 0.92 & 1.04 & 1.70 & 0.16 &0.36 & 0.58 & 0.70 & 1.08  \\
				\hline
				Ours (hybrid) & 0.41 & 0.63 & 1.07 & 1.14 & 2.25 & \textbf{0.24} & \textbf{0.40} & \textbf{0.75} & \textbf{0.96} & \textbf{1.47} & \textbf{0.27} & \textbf{0.58} & \textbf{0.85} & \textbf{0.97} & \textbf{1.58} & \textbf{0.14} & \textbf{0.32} & \textbf{0.53} & \textbf{0.64} & \textbf{0.91}\\
				\hline
				\hline
				
				Motion &  \multicolumn{5}{c}{Waiting} 
				& \multicolumn{5}{|c}{Walking Dog} & \multicolumn{5}{|c|}{Walking Together} & \multicolumn{5}{c}{Average}\\
				\hline
				millisecond & 80&160&320&400&1k & 80&160&320&400&1k & 80&160&320&400&1k & 80&160&320&400&1k  \\
				\hline
				Res-sup~\cite{martinez2017human}  & 0.32 & 0.63 & 1.07 & 1.26 &--& 0.52 & 0.89 & 1.25 & 1.40 &--& 0.27 & 0.53 & 0.74 & 0.79 &--& 0.40 & 0.69 & 1.04 & 1.18 & --\\
				CSM~\cite{li2018convolutional}  & 0.30 & 0.62 & 1.09 & 1.30 &2.50& 0.59 & 1.00 & 1.32 & 1.44 &1.92& 0.27 & 0.52 & 0.71 & 0.74 &1.28& 0.38 & 0.68 & 1.01 & 1.13 & 1.77\\
				Traj-GCN~\cite{liu2019towards}  & 0.23 & 0.50 & 0.92 & 1.15&-- & 0.46 & 0.80 & {1.12} & {1.30} &--& {0.15} & 0.35 & 0.52 & {0.57} &--& {0.27} & 0.53 & 0.85 & 0.96 & --\\
				DMGNN \cite{li2020dynamic} & {0.22} & {0.49} & {0.88} & {1.10} &--& {0.42} & {0.72} & {1.16} & {1.34}& --& {0.15} & {0.33} & {\bf 0.50} & {0.57} &--& {0.27} & {0.52} & {0.83} & {0.95} & --\\
				LTD-10-10 \cite{mao2019learning} & 0.23 & 0.50 & 0.91 & 1.14 & 2.37 & 0.46 & 0.79 & 1.12 & 1.29 & 1.86 & 0.15 & 0.34 & 0.52 & 0.57 & 1.16 & 0.27 & 0.52 & 0.83 & 0.95 & 1.62\\
				HisRepItself \cite{wei2020his}& 0.22 & 0.49 & 0.92 & 1.44 & 2.30 &0.46 & 0.78 & 1.05 & 1.23 & 1.82 & \textbf{0.14} & 0.32 & \textbf{0.50} & \textbf{0.55} & 1.16 & 0.27 & 0.52 & 0.82 &0.94 & 1.57 \\
				\hline
				Ours (hybrid) & \textbf{0.21} & \textbf{0.46} & \textbf{0.83} & \textbf{1.03} & \textbf{2.18} &\textbf{0.41} & \textbf{0.71} & \textbf{1.01} & \textbf{1.15} & \textbf{1.72} & \textbf{0.14} & \textbf{0.31} & \textbf{0.50} & 0.56 & \textbf{1.15} & \textbf{0.25} & \textbf{0.48} & \textbf{0.79} & \textbf{0.91} & \textbf{1.51}\\
				\hline
	\end{tabular}}}	\vspace{0.cm}
	\caption{Comparison of mean angle errors (MAE) of different methods for both short-term ($\leq$ 400 milliseconds) and long-term prediction (1k milliseconds) on the other 11 actions in the Human3.6M dataset. "--" indicates that the original paper did not report the result.}
	\label{tab:pred_h36m_11}
\end{table*}

\begin{figure*}[htbp]
	\centering
	\includegraphics[width=\textwidth]{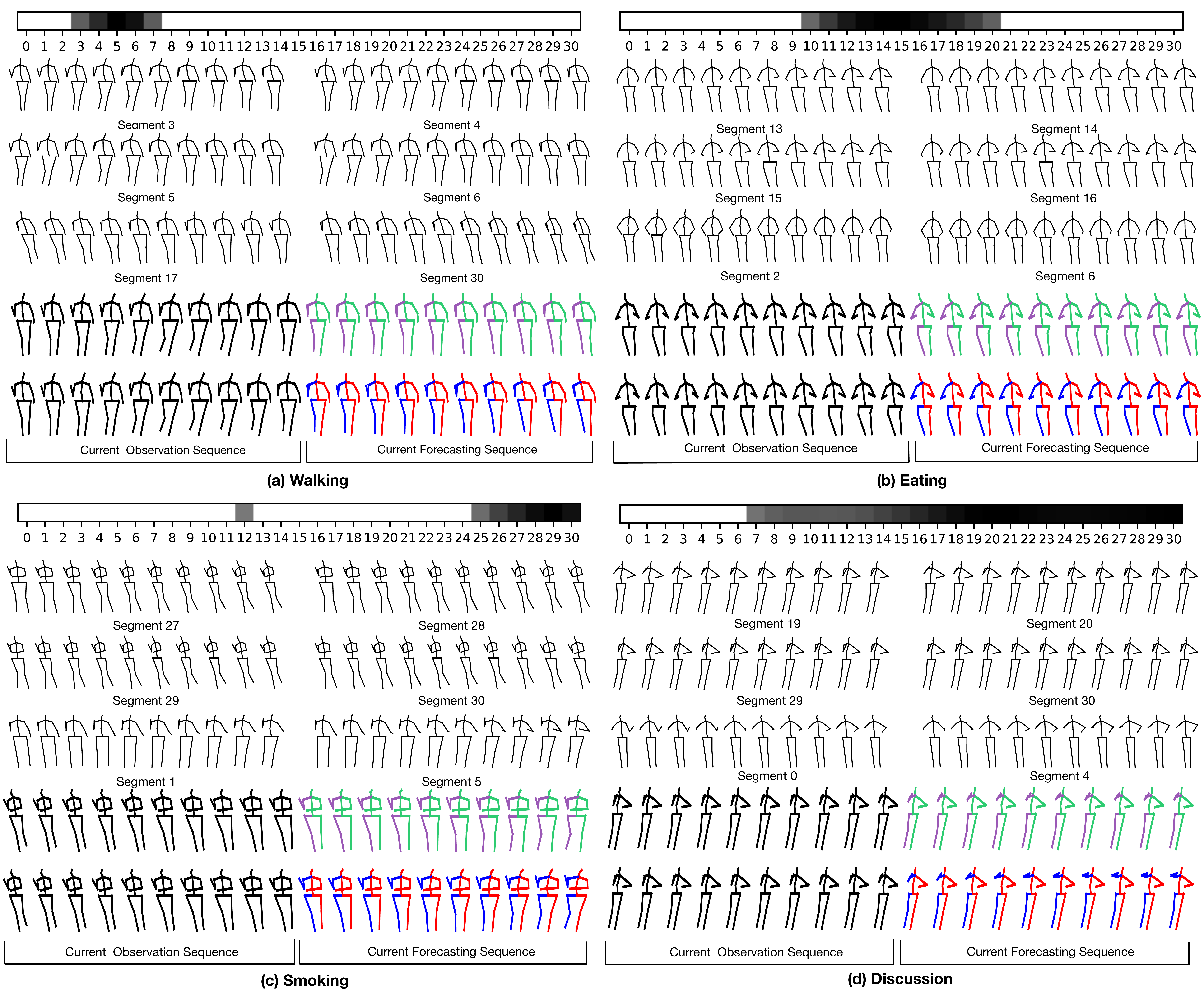}
	\caption{The visualization of human skeleton motion forecasting of four typical actions with hybrid attention maps. The black skeletons at the bottom are the latest observation sequences which are used to calculate current attention weights. The purple-green skeletons are the prediction hypotheses of our method. The blue-red skeletons are the ground truth. In our experimental setting, for each case there are 31 available history motion segments with a length of 10 frames for the RL hard attention module to select and the soft attention is only applied to the selected segments. In the hybrid attention maps, darker colors indicate larger attention weights. White color means the corresponding motion segment is not selected as key information. Best viewed in color.}
	\label{fig:skeleton_plot}
\end{figure*}

\subsubsection{Alternating Training Strategy}

The contextual encoding function is initialized with well-trained parameters in the pre-training stage and fixed in the formal-training stage.
The soft attention based motion generator is initialized with the pre-trained model in \cite{wei2020his}.
We perform alternating optimization of RL-HA and motion generator (MG) modules: (a) train the RL-HA module with fixed parameters of MG with Double Deep Q-Learning (DDQN) method \cite{hasselt2016deep}; (b) finetune MG with fixed parameters of RL-HA using back-propagation methods.

\section{Additional Experimental Results}

In this section, we provide complementary experimental results on the nuScenes and Human3.6M datasets.

\begin{table}[!tbp]
	\centering
	\setlength{\tabcolsep}{3mm}{
		\resizebox{\columnwidth}{!}{
			\begin{tabular}{l|ccc}
				\hline
				& full+soft &  hybrid (w/o GMP) & hybrid (w GMP) \\
				\hline 
				Vehicle & 0.63/1.28 & 0.59/1.23 & \textbf{0.54}/\textbf{1.12}  \\
				Pedestrian & 0.32/0.56 & 0.31/0.54 &  \textbf{0.26}/\textbf{0.51}  \\
				\hline
	\end{tabular}}}
	\vspace{0.0cm}
	\caption{4.0s minADE$_{20}$/minFDE$_{20}$ in different ablation settings (nuScenes).}\label{tab:nuScenes_GMP} 
	\vspace{-0.2cm}
\end{table}

\begin{table}[!tbp]
	\centering
	\setlength{\tabcolsep}{3.5mm}{
		\resizebox{\columnwidth}{!}{
			\begin{tabular}{l|cccc}
				\hline
				& $\tau=2$ & $\tau=5$ & $\bm{\tau=10}$ & $\tau=20$ \\
				\hline 
				Vehicle & 0.50/1.08 & 0.52/1.09 & 0.54/1.12 & 0.58/1.17 \\
				Pedestrian & 0.24/0.50 & 0.25/0.51 & 0.26/0.51 & 0.29/0.54  \\
				\hline
	\end{tabular}}}
	\vspace{0.0cm}
	\caption{4.0s minADE$_{20}$/minFDE$_{20}$ with different $\tau$ values (nuScenes).}\label{tab:nuScenes_tau} 
	\vspace{-0.5cm}
\end{table}

\subsection{nuScenes Dataset: Traffic Scenarios}
Besides the ablation studies (Table \ref{tab:nuScenes_veh}, \ref{tab:nuScenes_ped}) in the main paper, we show the results of an additional ablation setting hybrid (w/o GMP) in Table \ref{tab:nuScenes_GMP} to illustrate the effectiveness of GMP module. In this setting, the RL agent can only observe the self features of each node (agent) without knowing the relational/social features extracted by GMP, which leads to larger prediction errors compared to hybrid (w/ GMP).
Note that hybrid (w/o GMP) outperforms full+soft, which implies the RL based hybrid attention can still help with motion prediction even without relational features. These results demonstrate that all modules are indispensable with improvement. 

In our experiments, we used $\tau=10$ (10 frames in 2.0s). We show additional results with different $\tau$ values in Table \ref{tab:nuScenes_tau}.
Generally, the minADE$_{20}$/minFDE$_{20}$ reduces as $\tau$ becomes smaller (i.e., the frequency of graph topology inference increases), which implies the advantage of dynamic prediction mechanism. 
However, the running time will increase as $\tau$ becomes smaller.

\subsection{Human3.6M Dataset: Human Motions}
We provide the comparison of prediction results on the other 11 actions for skeleton based human motion forecasting in Table \ref{tab:pred_h36m_11}.
It is shown that \textit{Ours (hybrid)} achieves the smallest MAE in most actions as well as in average compared with baselines.
The action ``Directions'' is an interesting exception where HisRepItself outperforms our method.
A potential reason is that in the ``Directions'' action, there is no clear pattern of the temporal dependency between the current observation and previous motions, which makes it hard for the RL-agent to discriminate and select the appropriate history motion segments to pay attention to.

We also visualize the prediction of human skeletons and the learned hybrid attention weights in typical testing cases in Figure \ref{fig:skeleton_plot}.
It shows that our method can accurately forecast the human motions. 
More specifically, we visualize the top four motion segments with the largest four hybrid attention weights in each case (i.e., the motion segments in the first two rows). It shows that these segments have similar patterns with the current observation sequence and thus are selected as key information, which implies that the learned hybrid attention is reasonable and interpretable. 
We also visualize some irrelevant segments that are discarded by the model (i.e., the motion segments in the third row). It can be easily found that these segments are dissimilar to the latest observation sequence, thus are unimportant for the current prediction. 

In addition, in case (a), (b) and (c), the learned hybrid attention is sparse, which implies that the model is able to effectively discriminate and only focus on the key information. An interesting exception is case (d), where most history segments are selected. A potential reason is that most history segments in this case are very similar to the current observation sequence, which leads the model to take them all into account for current prediction.

\section{Further Experimental Details}

In this section, we provide further details of the experiments, which includes dataset generation, baseline methods, as well as implementation details.

\subsection{Datasets and Evaluation Metrics}
\subsubsection{Mixed Particle Simulation}
In the mixed particle system, there are two types of particles: charged particles and uncharged particles. The charged particles are uniformly sampled to carry positive or negative charges $q_i \in \{\pm q\}$, which interact with each other via Coulomb forces, which is given by
\begin{equation}
	F_{ij} = C \cdot (q_i \cdot q_j) \cdot \frac{r_i-r_j}{||r_i-r_j||^3},
\end{equation}
where $C$ is a constant. These charged particles may either attract or repel each other, although the forces may be weak when the distance in between is large.
However, the motions of uncharged particles are totally independent since there is no force applied to them. They move straight with a constant velocity the same as the random initialization. In this paper, we have 3 charged particles and 3 non-charged ones in each case. We generated 8k samples for training, and 4k samples for validation and testing, respectively.

The simulation process of charged particles is mainly adopted from NRI \cite{kipf2018neural}. In order to prevent the force divergence issue when two particles move with a very small distance, we adopt the same strategy as suggested in \cite{kipf2018neural} to avoid numerical issues, which is to clip the forces to some maximum threshold. Despite that this is not exactly physically precise, the generated trajectories are not distinguishable to human observers and do not affect the conclusion of the paper.

The evaluation metric for trajectory prediction in this experiment is the mean squared error, which is calculated by
\begin{equation}
	\text{MSE} = \frac{1}{NT_f} \sum_{i=1}^{N} \sum_{t'=t+1}^{t+T_f}||\mathbf{x}^i_{t'} - \hat{\mathbf{x}}^i_{t'}||^2.
\end{equation}

\subsubsection{nuScenes Dataset \cite{nuscenes2019}} 

The nuScenes dataset is a widely used large-scale driving dataset with a full set of sensor suite, which was collected in Boston and Singapore. It provides the point cloud information, trajectory annotations of heterogeneous traffic participants (e.g., cars, pedestrians and cyclists), as well as the map information.
We processed the original data into segments with a length of 6 seconds to construct our dataset (2 seconds as history and 4 seconds as future). We generated about 8k samples for training, and 2k samples for validation and testing, respectively.

We adopt the standard evaluation metrics in trajectory prediction, which include minADE$_K$, minFDE$_K$ and miss rate (MR). In this paper, we use the same $K=20$ as most baselines. 
The MR(@$d$m) is calculated by
{\small
	\begin{align}
		\text{MR(@}d\text{m}) = \frac{1}{N} \sum_{i=1}^{N} \mathbb{I}\left(\min_k ||\mathbf{x}^{i,k}_{t+T_f} - \hat{\mathbf{x}}^{i,k}_{t+T_f}||_2 > d\right),
\end{align}}
where $\mathbb{I}(\cdot)$ denotes an indicator function to indicate whether the current case is a failure case, and $d$ is a manually defined threshold.

\subsubsection{Human3.6M Dataset \cite{h36m_pami}}
The Human3.6M dataset is a widely used skeleton-based human motion dataset for pose estimation and motion forecasting, which includes 15 different activities performed by 7 professional actors.
The human skeleton information is provided in two representations: 3D joint positions and joint angles. 
The skeleton has 32 joints, the 3D coordinates of which can be computed by applying the forward kinematics. As in \cite{wei2020his}, we also down-sample the motion sequences to 25 frames per second, and remove the global rotation, translation and constant angles. 
In this paper, we chose relative angles between joint to represent the skeleton state.

\subsection{Baseline Methods}

\subsubsection{Ablative Baselines}
\begin{itemize}
	\item Ours (true+soft): This is the model that only applies soft graph attention to the true relation graph. Note that this model is only used for the experiments on mixed particle simulation since the true relation graph is not accessible in real-world scenarios and dataset.
	\item Ours (full+soft): This is the model that only applies soft graph attention to a fully connected relation graph.
	\item Ours (ELBO+soft): This is the model where only the RL-HA module is replaced by an ELBO based module with other modules not changed, which is trained end-to-end. The purpose of this ablation setting is to provide a baseline based on an alternative way to obtain hard attention.
	\item Ours (hybrid, static): This is the model that applies both RL hard attention to obtain the inferred relation graph and soft attention to figure out relative importance. The inferred relation graph remains static during the whole prediction horizon and the model performs one-shot prediction.
	\item Ours (hybrid, dynamic): This model setup is very similar to Ours (hybrid, static). The difference is that the model performs iterative prediction with a fixed horizon of sliding window. The inferred relation graph is dynamically evolving over time.
	\item Ours (hybrid): This model setup is only used for human motion prediction. The RL hybrid attention and soft attention work together to extract informative history features for the motion generator in \cite{wei2020his}.
\end{itemize}

\subsubsection{For Mixed Particle Simulation}

\begin{itemize}
	\item Corr. (LSTM): The baseline method for edge recognition used in \cite{kipf2018neural}.
	\item LSTM (single) / LSTM (joint): The baseline methods for state sequence prediction in \cite{kipf2018neural}.
	\item NRI: The NRI model with static latent interaction graph \cite{kipf2018neural}.
	\item DNRI: A model for neural relational inference with dynamic interaction graphs \cite{graber2020dynamic}.
	\item Supervised: Since the true relation graph is accessible in the simulation data, we can use supervised learning to train a binary classifier to infer the existence of the edges in the graph. The ground truth labels include 0 (w/o edge) and 1 (w/ edge).
\end{itemize}

\subsection{Implementation Details}\label{sec:implementation_detail}
In this section, we introduce the details of model architecture, hyperparameters and specific experimental settings for each dataset.

\subsubsection{For Multi-Agent Interacting Systems}
\vspace{-0.1cm}
We trained the models for 100 epochs for both particle simulation and nuScenes dataset. They shared the same model architecture and specific details of model components are introduced below:
\begin{itemize}
	\item GMP: The state embedding functions $f^m_s$, $f^m_n$, node attribute update function $f_v$, and the encoding function $f_\text{enc}$ are all three-layer MLPs with hidden size = 64. During the pre-training stage, the decoding function is also a three-layer MLP with hidden size = 64. The context embedding function $f_c$ is a four-layer convolutional block with kernel size = 5. The layer structure is [[Conv, ReLU, Conv, ReLU, Pool], [Conv, ReLU, Conv, ReLU, Pool]]. The context embedding is only applied to the ``traffic scenario'', where the context information is the projected point cloud images.
	\item RL-HA: The maximum RL-step $\eta$ is set to 10. In the total reward, the hyperparameters are $\beta_{\text{imp}} = \beta_{\text{sti}} = \beta_{\text{pun}} =0.01$. We also set $\Omega_s = \Omega_p = 1.0$. 
	In the ``traffic scenario'', we define a ``success case'' where the end-point error is less than the miss rate threshold and a ``failure case'' as the opposite. The stimulation reward is applied when the current case changes from ``failure case'' to ``success case'', and the punishment reward is applied for the opposite situation.
	
	All the coefficients and thresholds in reward function were decided empirically. 
	$R_{\text{reg}}$/$R_{\text{imp}}$ reward the overall improvement of prediction while $R_{\text{sti}}$/$R_{\text{pun}}$ are mainly related to endpoint prediction.
	We found that increasing the weights of $R_{\text{sti}}$/$R_{\text{pun}}$ could improve both minADE and minFDE in a certain range while overly large weights could have negative effects on minADE. 
	The miss rate thresholds should be specifically decided for various types of agents. 
	
	\item SGA-MG: The Embedding LSTM (E-LSTM) and Generation LSTM (G-LSTM) both have a hidden size of 128. For the particle simulation, we performed one-shot prediction with a static inferred relation graph; for the nuScenes dataset, we performed progressive forecasting with a sliding window of 2 seconds (10 frames) with dynamic relation graphs.
\end{itemize}

\vspace{-0.3cm}
\subsubsection{For Human Skeleton Motions}\label{sec:human_details}
\vspace{-0.1cm}
We trained the models for 50 epochs on the Human3.6M dataset. 
We adopted exactly the same experimental settings as \cite{wei2020his}. More specifically,
during training, we trained the model to predict the future 10 frames based on the history 50 frames and the attention weights are calculated based on the latest observation sequence with 10 frames.
During testing, we enabled progressive long-term prediction with a sliding window to generate future 25 frames.

Specific details of model components are introduced in the following:
\begin{itemize}
	\item Encoding / Decoding (pre-training stage): They are three-layer MLPs with hidden size = 128.
	\item RL-HA: The maximum RL-step $\eta$ is set to 10. In the total reward, the hyperparameter is $\beta_{\text{imp}} = 0.01$.
	\item Motion Generator: We adopted the same model architecture and hyperparameters as in \cite{wei2020his}.
\end{itemize}

\end{document}